\documentclass{article}

\PassOptionsToPackage{numbers}{natbib}

\usepackage[preprint]{sty_neurips_2024}

\usepackage[utf8]{inputenc} 
\usepackage[T1]{fontenc}    
\usepackage{hyperref}       
\usepackage{url}            
\usepackage{booktabs}       
\usepackage{amsfonts}       
\usepackage{nicefrac}       
\usepackage{microtype}      
\usepackage{xcolor}         

\usepackage{sty_abbrv}
\usepackage{tabularx}
\usepackage{arydshln}
\usepackage{multirow}
\newcolumntype{Y}{>{\centering\arraybackslash}X}
\usepackage{float}
\usepackage{cleveref}
\usepackage{graphicx}
\graphicspath{{images/}}
\usepackage{xurl}
\usepackage{array}
\usepackage{wrapfig}
\usepackage{subcaption}

\makeatletter
\newcommand{\thickhline}{%
	\noalign {\ifnum 0=`}\fi \hrule height 1pt
	\futurelet \reserved@a \@xhline
}
\newcolumntype{"}{@{\hskip\tabcolsep\vrule width 1pt\hskip\tabcolsep}}

\def\Cline#1#2{\@Cline#1#2\@nil}
\def\@Cline#1-#2#3\@nil{%
	\omit
	\@multicnt#1%
	\advance\@multispan\m@ne
	\ifnum\@multicnt=\@ne\@firstofone{&\omit}\fi
	\@multicnt#2%
	\advance\@multicnt-#1%
	\advance\@multispan\@ne
	\leaders\hrule\@height#3\hfill
	\cr}
\makeatother

\usepackage{pifont}
\newcommand{\cmark}{\ding{51}}%
\newcommand{\xmark}{\ding{55}}%

\usepackage{standalone}
\usepackage{wrapfig}

\title{Understanding the Cross-Domain Capabilities of Video-Based Few-Shot Action Recognition Models}

\author{%
	Georgia Markham\thanks{Corresponding Author. \texttt{georgia.markham@sydney.edu.au }} \;\;\;\;\;\;\;\;\;\;\;\;  Mehala Balamurali \;\;\;\;\;\;\;\;\;\;\;\;  Andrew J. Hill \\[0.3cm]
	Australian Centre for Robotics\\
	The University of Sydney
}

\begin{document}

\maketitle

\begin{abstract}
  Few-shot action recognition (FSAR) aims to learn a model capable of identifying novel actions in videos using only a few examples. In assuming the base dataset seen during meta-training and novel dataset used for evaluation can come from different domains, cross-domain few-shot learning alleviates data collection and annotation costs required by methods with greater supervision and conventional (single-domain) few-shot methods. While this form of learning has been extensively studied for image classification, studies in cross-domain FSAR (CD-FSAR) are limited to proposing a model, rather than first understanding the cross-domain capabilities of existing models. To this end, we systematically evaluate existing state-of-the-art single-domain, transfer-based, and cross-domain FSAR methods on new cross-domain tasks with increasing difficulty, measured based on the domain shift between the base and novel set. Our empirical meta-analysis reveals a correlation between domain difference and downstream few-shot performance, and uncovers several important insights into which model aspects are effective for CD-FSAR and which need further development. Namely, we find that as the domain difference increases, the simple transfer-learning approach outperforms other methods by over 12 percentage points, and under these more challenging cross-domain settings, the specialised cross-domain model achieves the lowest performance. We also witness state-of-the-art single-domain FSAR models which use temporal alignment achieving similar or worse performance than earlier methods which do not, suggesting existing temporal alignment techniques fail to generalise on unseen domains. To the best of our knowledge, we are the first to systematically study the CD-FSAR problem in-depth. We hope the insights and challenges revealed in our study inspires and informs future work in these directions.

\end{abstract}

\section{Introduction}

The ability to recognise video contents is crucial to many real-world applications including human-robot interaction \cite{reddy_synthetic--real_2023}, surveillance \cite{arunnehru_human_2018}, anomaly and hazard detection \cite{li_computer_2022}, and performance assessment and monitoring \cite{martin_automatic_2023}. Following advancements made in deep-learning, a large focus has been on developing video action recognition methods which are trained under full supervision. While these methods perform well, a large amount of labelled examples per action class are required to achieve this performance. This labour intensive human annotation and data collection requirement restricts the real-world usability of these methods, especially in scenarios where it can be difficult to collect a sufficient amount of examples per class, or deployment time is critical.

Alleviating these concerns, few-shot learning (FSL) aims to learn a model capable of identifying novel classes from only a few examples of each. Under this setting, FSL models are trained on a set of \textit{base} classes and evaluated on a class-wise disjoint set of \textit{novel} classes.  While FSL is notionally more suited for real-world use, recent developments have been made under the assumption that the base and novel classes are sampled from the same domain (single-domain). This protocol inadvertently requires the collection and annotation of \textit{similar} base class data, and cannot be applied to datasets containing a small number of classes, such as the recently released RoCoG-v2 (Robot Control Gestures) dataset \cite{reddy_synthetic--real_2023}, containing only 7 actions for commanding a drone. 

\begin{wrapfigure}{R}{0.48\linewidth}
	\vspace{-0.2cm}
	\centering
	\includegraphics[width=\linewidth]{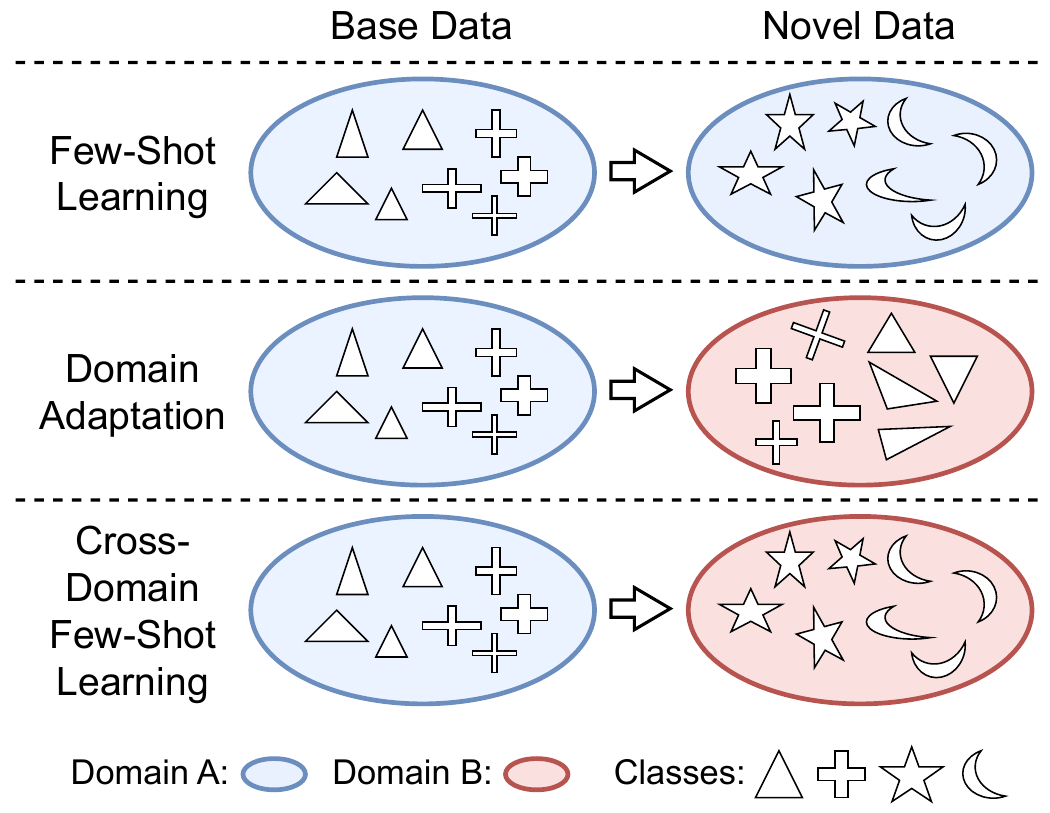}
	\caption{An illustrative comparison between single-domain FSL, domain adaptation, and cross-domain FSL. This paper concerns cross-domain FSL for action recognition in video.}
	\label{fig:example-cross-domain}
\end{wrapfigure}

Alleviating these concerns further, cross-domain few-shot learning (CD-FSL) assumes the base and novel classes are sampled from different domains (see \Cref{fig:example-cross-domain} for an illustrative comparison). The proposal of the more challenging and realistic BSCD-FSL benchmark \cite{vedaldi_broader_2020}, consisting of four diverse image datasets, facilitated key insights into which types of few-shot image classification methods are most effective in practice to be drawn. This ultimately drove advancements in few-shot image classification towards real-world usability. However, while CD-FSL has been explored for image classification \cite{chen_closer_2019, triantafillou_meta-dataset_2020, vedaldi_broader_2020, luo_closer_2023, wang_cross-domain_2021, xu_memrein_2022, li_cross-domain_2022}, it remains widely unexplored for video-based action recognition. To date, only two specific cross-domain few-shot action recognition (CD-FSAR) models have been proposed, however little is understood about the capabilities of existing models under cross-domain settings, which could inform future developments in this field. 

To this end, we conduct a systematic empirical study to explore and understand the challenges of CD-FSAR. To do so, we first analyse and select cross-domain settings (base and novel dataset combinations) to evaluate models on. We select settings with a range of difficulties, where we define a setting as `more challenging' when there is a larger measure in domain shift. Here, the Maximum Mean Discrepancy (MMD) \cite{gretton_kernel_2012} score is used. We select from datasets with varying degrees of similarity in terms of their spatial and temporal characteristics, including the commonly used (HMDB51 \cite{kuehne_hmdb_2011}, UCF101 \cite{soomro_ucf101_2012}, and SSv2 \cite{goyal_something_2017}), less commonly used (Diving48 \cite{ferrari_resound_2018}), and never before used (RoCoGv2 \cite{reddy_synthetic--real_2023}) datasets for FSAR. We then re-evaluate a diverse set of existing state-of-the-art (SOTA) single-domain, transfer-based, and cross-domain FSAR models under these new cross-domain settings, to analyse their effectiveness across settings with varying difficulty. 

Our results demonstrate that as the domain difference increases, the simple transfer-learning based method outperforms the single-domain and cross-domain methods by over 12 percentage points. Under these new, more challenging settings, the specialised cross-domain method achieves the lowest performance, comparable to random weight model performances. The SOTA single-domain FSAR models which use temporal alignment achieve similar or worse performance than earlier methods which omit temporal alignment, suggesting existing temporal alignment techniques fail to generalise on unseen domains. Lastly, we witness a correlation between domain difference and downstream few-shot performance, suggesting this simple measure could be used to inform base dataset design. However, in some instances adding more data to the base dataset boosts few-shot performance, whereas in others it reduces, revealing a sensitivity to base class construction. We hope the insights and challenges revealed in our study inspires and informs future work in these directions.

\vspace{-0.1cm}
\section{Related Work}

\vspace{-0.1cm}
\subsection{Few-Shot Learning (FSL)}

Inspired by human ability, FSL aims to identify novel classes using only a few labelled examples of each. Early, well established approaches follow the `learning to learn' meta-learning paradigm, whereby models are trained \textit{episodically} to emulate the evaluation setting. Of these, metric-learning based approaches such as ProtoNet \cite{snell_prototypical_2017} and MatchingNet \cite{vinyals_matching_2016} assume that features discriminative of the base classes are also discriminative of novel classes, and thus learn an encoder and similarity function based on Euclidean distance or cosine distance in feature space, for example. In contrast, transfer-based approaches \cite{chen_closer_2019} train an encoder with linear classification head using all data from the base dataset, then, upon evaluation, the classification head gets replaced and fine-tuned using the few labelled examples. Models specific to FSAR measure temporal alignment between sequences of frames in feature space to account for the challenging addition of the temporal dimension within videos. For example, STRM \cite{thatipelli_spatio-temporal_2022} encodes patch-level (spatio) and frame-level (temporal) information into features and leverages an attention-based mechanism developed by \cite{perrett_temporal-relational_2021} to align query and support set sub-sequences. MoLo \cite{wang_molo_2023} designs a long-short contrastive objective to improve temporal context awareness during the matching process. Despite the progress shown in this field, the majority of FSAR models assume the base and novel datasets come from the same domain (termed single-domain), and thus may fail to generalise to unseen, potentially out-of-distribution, classes.

\subsection{Cross-Domain Few-Shot Learning (CD-FSL)}

In CD-FSL, the base and novel class sets can be drawn from different domains, which is indicative of how few-shot models would be utilised in the real-world. Chen \etal \cite{chen_closer_2019} were the first to highlight the domain difference problem by studying the cross-domain generalisation ability of existing few-shot image classification models, finding that they did not generalise well to the unseen domains. To foster progress in this field, Triantafillou \etal \cite{triantafillou_meta-dataset_2020} proposed the META-DATASET benchmark which prescribes the use of 10 diverse datasets for evaluating models under multi-domain (multiple datasets in both the base and novel sets, with shared domains) and cross-domain settings. To then shift CD-FSL research toward real-world usability, Guo \etal \cite{vedaldi_broader_2020} proposed a purely cross-domain benchmark, BSCD-FSL. This benchmark uses only ImageNet as the base dataset, and varying real-world image datasets which qualitatively have increasing dissimilarity to natural images as novel datasets. Using this more challenging benchmark, Guo \etal discovered that existing single-domain and cross-domain methods were outperformed by simple transfer-based approaches. Analogous to the work of Guo \etal, we re-evaluate existing FSL methods under more challenging cross-domain scenarios, however we study video-based action recognition models rather than image classification models, and we justify our cross-domain scenario choices using a quantified measure of similarity, rather than a qualitative perceived similarity, to ensure a range of domain difference difficulties are tested and trends and correlations can be more reliably inferred. 
 
Following the proposal of more challenging cross-domain benchmarks for few-shot image classification, subsequent works developed methods that perform better under these more realistic settings  \cite{tseng_cross-domain_2020, wang_cross-domain_2021, fu_meta-fdmixup_2022, xu_memrein_2022, li_cross-domain_2022,perera_discriminative_2024, liang_boosting_2021,  phoo_self-training_2020, islam_dynamic_2021}. However, in the field of CD-FSAR, only two methods have been developed \cite{wang_cross-domain_2023, samarasinghe_cdfsl-v_2023} and a challenging benchmark does not exist. Inspired by the few-shot image classification works of Phoo and Hariharan \cite{phoo_self-training_2020} and Islam \etal \cite{islam_dynamic_2021}, the two existing CD-FSAR methods include a self-supervised pre-training stage, leveraging unlabelled data from both the base and novel sets. These methods hence assume a substantial amount of unlabelled novel data is available at training time. In practice, this is an unrealistic assumption which implicitly requires the collection and trimming of videos from the intended novel domain, which we argue is a form of labelling, and thus violates the few-shot assumption. Our study takes a step back to understand the capabilities of a range of existing FSAR methods under diverse cross-domain scenarios, to inform where future efforts in this field should be focused. We are the first to systematically study this problem in-depth.

\subsection{Impact of Base Dataset Design}

Base dataset design refers to characteristics of the dataset that few-shot models learn from during training (the base dataset). This can include the number of classes overall, number of examples per class, and similarity to the novel dataset, among others. Sbai \etal \cite{vedaldi_impact_2020} were one of the first to systematically study how certain characteristics, like the aforementioned, affect the resulting features used in few-shot image classification. Notably, they found that similarity of the base classes and novel classes has a crucial effect. This work however experiments with only two image datasets and they do not use a quantified measure of similarity, they simply state a combination is more similar when the base and novel sets come from the same dataset. Complementing this, Luo \etal \cite{luo_closer_2023} comprehensively study various aspects of the scale of the training dataset, with a key insight revealing that larger datasets, both in terms of increasing the number of classes and samples per class, may lead to degraded performance on certain downstream datasets. This insight contradicts the common finding in machine learning where learning from more data often leads to better generalisation. No base dataset design study exists for FSAR. The addition of the temporal dimension may lead to findings differing from those discovered with image data. Our study acts as a preliminary investigation.

\section{Cross-Domain Few-Shot Task Formulation}

Few-shot classification aims to learn a model that can identify novel classes using only a few examples of each. More formally, when given a data point from the \textit{query} set $\mathcal{Q}$, a few-shot classification model is tasked with classifying it as one of \textit{N} classes, based on \textit{K} examples of each class available within the \textit{support} set $\mathcal{S}$. Tasks constructed in this way are known as \textit{episodes} which can be summarised as \textit{N}-way \textit{K}-shot tasks. To \textit{learn} a model capable of doing so, a model first undergoes meta-training using a \textit{base} set of classes, $\mathcal{D}_{base}$. Then, at test time, \textit{N}-way \textit{K}-shot episodes are constructed by sampling from a novel set of classes, $\mathcal{D}_{novel}$, to create both $\mathcal{Q}$ and $\mathcal{S}$, and classification performance on $\mathcal{S}$ is calculated. Note that $\mathcal{D}_{base}$ and $\mathcal{D}_{novel}$ are disjoint class-wise.

Conventionally, the classes within $\mathcal{D}_{base}$ and $\mathcal{D}_{novel}$ are sampled from a single dataset (i.e., single-domain). However, under a cross-domain setting, $\mathcal{D}_{base}$ and $\mathcal{D}_{novel}$ are constructed from different datasets. Additionally, we propose that $\mathcal{D}_{base}$ can consist of many datasets (still independent of $\mathcal{D}_{novel}$), such that $\mathcal{D}_{base} = D_1\cup D_2\cup ...\cup D_n$. This differs from the conditions previously used to evaluate CD-FSAR models \cite{wang_cross-domain_2023, samarasinghe_cdfsl-v_2023}, where $\mathcal{D}_{base}$ consisted of a single dataset.

\section{Dataset Analysis}\label{sec:dataset-analysis}

Since no datasets or benchmarks dedicated to CD-FSAR exist, we systematically select combinations from five existing action recognition datasets. In summary, we justify our choices based on the measure of domain difference between the base and novel sets, choosing combinations which provide a range of measures to ensure the cross-domain capabilities of the re-evaluated models are widely tested. We re-evaluate models on the chosen combinations in \Cref{sec:results}.

\subsection{Datasets}\label{sec:datasets}

We endeavoured to analyse datasets which span a variety of human action concepts (full-body motion in everyday activities, full-body motion in sports, gestures, and human-object interactions) and thus qualitatively have varying degrees of similarity in terms of their spatial and temporal characteristics. Here, we provide an overview of five publicly available action recognition datasets which we will measure domain difference between and then select from. Refer to \Cref{tab:dataset-summary} for a summary of numeric statistics. Refer to \Cref{app:dataset_examples} to see frames of example videos sampled from each dataset, which exemplifies their qualitative differences. 

\paragraph{HMDB51.} \cite{kuehne_hmdb_2011} The Human-Motion Database is comprised of everyday human actions, such as `kiss', `jump', and `cartwheel', in videos from various sources, including movies and the internet, and thus has a diverse distribution with different backgrounds, camera angles, and ranges of motion.

\paragraph{UCF101.} \cite{soomro_ucf101_2012} Collected from YouTube, UCF101 contains action categories involving human-object interaction, body-motion only, human-human interaction, playing musical instruments, and sports. Therefore, this dataset has the largest diversity of actions.

\paragraph{SSv2.} \cite{goyal_something_2017} The Something Something video database contains fine-grained categories of human-object interaction scenarios which take place in everyday tasks. Classes typically involve doing `something' to `something', and thus require strong temporal reasoning. The video perspectives often have only hands within the frame. In this study we use the SSv2-Full train/val/test splits from \cite{zhu_compound_2018}. 

HMDB51, UCF101, and SSv2 are the most commonly used datasets for single-domain FSAR, and have also been used as the novel dataset for CD-FSAR \cite{samarasinghe_cdfsl-v_2023, wang_cross-domain_2023}. 
\begin{table}[H]
	\centering
	\renewcommand{\arraystretch}{1.1}
	\caption{Summary of the class and video composition of each dataset. See \Cref{app:dataset-stats} for further statistics.}
		
	\begin{tabularx}{\textwidth}{l *{5}{Y} }
			
			\Cline{2-6}{0.08em}
			
			
			& HMDB51
			& UCF101
			& SSv2{\scriptsize (-Full)}
			& Diving48 
			& RoCoGv2   \\
			
			
			\toprule
			
			
			No. Videos Overall
			& 6766
			&  13320
			&   71796 
			&  16997
			&  178  \\
			
			
			No. Classes Overall
			& 51
			&  101
			&   100
			&  47
			&  7  \\
			
			
%
%
%
%
			
			
			Train/Val/Test Split Source
			& \cite{cao_few-shot_2020}
			&  \cite{cao_few-shot_2020}
			&  \cite{zhu_compound_2018}
			&  This paper
			&  -  \\
			
			
			\bottomrule
			
		\end{tabularx}
		
		\label{tab:dataset-summary}
		\vspace{-0.2cm}
\end{table}

\paragraph{Diving48.} \cite{ferrari_resound_2018} Consisting of fine-grained competitive diving actions, this dataset eliminates static background biases such as objects, scenes, and people, which exist in other datasets and can make action classes more easily identifiable. Thus, this dataset requires finer-grained temporal understanding due to the similarities between classes and lack of background bias. Since no class-wise train/val/test splits for this dataset have been proposed before, we create splits. The classes contained within each split are itemised in \Cref{app:diving_splits}, as well as the procedure followed to obtain them.

\paragraph{RoCoGv2.} \cite{reddy_synthetic--real_2023} The Robot Control Gestures dataset consists of humans performing arm motion gestures intended for commanding a drone. Recorded from an aerial perspective with a plain grass background, this dataset also eliminates background biases. Unlike the aforementioned datasets which contain conventional human actions, these arm motion gestures are specific to the application. While this dataset was created for studying synthetic-to-real domain adaptation, we use it to evaluate FSAR models for the first time. In this study, we use all real (as opposed to synthetic) cropped aerial videos. This dataset is 38x smaller than the next smallest dataset, and contains the smallest number of actions, however we believe this is a representative example of an application that FSL would be ideal for in the real-world. Since this dataset is relatively small, we use it for the novel dataset only.

\subsection{Measuring Domain Difference} \label{sec:measure}

For video datasets, the difference between two domains can be influenced by a number of factors, including the camera perspective, background content, human body content (e.g., full body vs. only hands), range of motion, interactions with objects, lighting, etc. Thus, explicitly measuring the difference between two distributions is challenging, especially for video data. A simple, but commonly used metric for measuring the difference between two distributions is the Maximum Mean Discrepancy (MMD) \cite{gretton_kernel_2012, wang_cross-domain_2023}. This is measured as the Euclidean distance between the mean value of the samples on one domain to the other in feature space.

To calculate the domain difference between two distributions (i.e., a base and novel dataset combination), similar to the procedure carried out in \cite{wang_cross-domain_2023}, we first obtain the feature space representation of each video by extracting features from 6 uniformly sampled frames using ResNet34 \cite{he_deep_2016} pretrained on ImageNet \cite{deng_imagenet_2009}, then taking the average. Next, 256 samples from each domain are randomly selected and the MMD between the two are calculated. This sampling step is repeated 100 times to obtain the average MMD. A larger MMD indicates a larger difference in domain. When using a dataset within the base set and novel set, only the training and testing split samples, respectively, were used, and when the base set consists of more than one dataset, their training splits are combined. 

\subsection{Choice of Cross-Domain Settings}\label{sec:choice}

\begin{figure}[H]
	\centering
	\includegraphics[width=\textwidth]{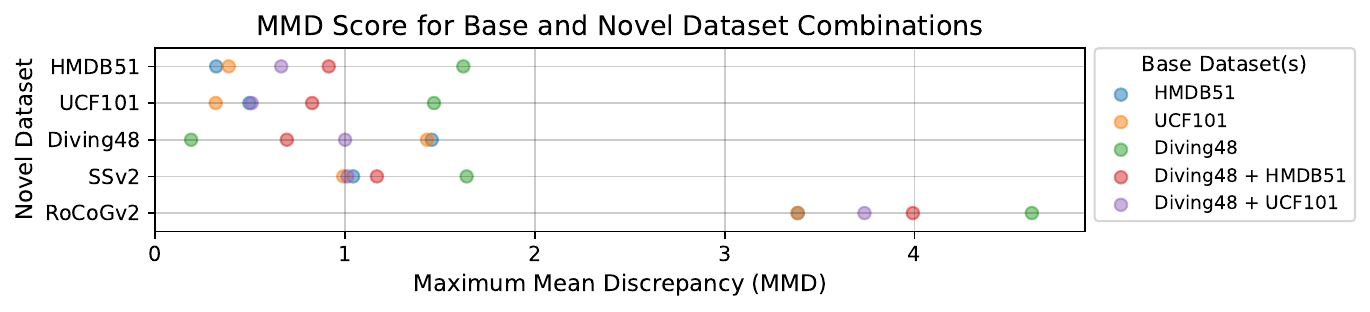}
	\caption{Domain differences for base dataset combinations used for evaluation in \Cref{sec:results}.}
	\label{fig:domain-difference}
\end{figure}

We allow the base dataset to consist of any number of the five datasets described in \Cref{sec:datasets}, except RoCoGv2 (since it is small), making 15 possible combinations, and use each of the five datasets as the novel set. Thus, following the above procedure, the domain difference for 75 combinations were calculated. The full results are shown in \Cref{app:domain-difference}. The subset of five base datasets chosen for model evaluation is shown in \Cref{fig:domain-difference}, as indicated by colour. We chose the subset that has the greatest range and cover of the MMD space. We note that among the datasets we studied, there were no combinations with an MMD between 2 and 3. 
 
\section{Evaluation Setup}\label{sec:eval-setup}

\subsection{Existing Models}\label{sec:models}

We re-evaluate 5 existing FSAR models, including ProtoNet \cite{snell_prototypical_2017}, STRM \cite{thatipelli_spatio-temporal_2022}, MoLo \cite{wang_molo_2023}, CDFSL-V \cite{samarasinghe_cdfsl-v_2023}, and a simple transfer-learning approach, dubbed \textit{Transfer} \cite{chen_closer_2019}. We believe these models are a representative set of well-established and SOTA models which have diverse characteristics. All models have publicly available code which were used for this study. We acknowledge this set is non-exhaustive, however it will provide us with preliminary insights. See \Cref{tab:model-comparison} for a comparative summary of notable model aspects, and see \Cref{app:models} for more detailed descriptions.

\begin{table}[H]
	\centering
	\caption{Summary of the existing few-shot action recognition models re-evaluated in this study.}
	
	\begin{tabularx}{\textwidth}{l *{3}{Y}c}
		
		\Cline{2-5}{0.08em}
		
		
		& Designed Scenario
		& Temporal Alignment?
		& Novel Domain Unseen?
		& Classification Strategy  \\
		
		
		\toprule
		
		
		ProtoNet \cite{snell_prototypical_2017}
		& single-domain
		&  \xmark
		&   \cmark
		&  Closest prototype \\
		
		
		STRM \cite{thatipelli_spatio-temporal_2022}
		&  single-domain
		&  \cmark
		&  \cmark
		&  Closest prototype  \\
		
		
		MoLo \cite{wang_molo_2023}
		&  single-domain
		&  \cmark
		&  \cmark
		&  Closest prototype  \\
		
		
		\textit{Transfer} \cite{chen_closer_2019} 
		&  -
		&  \xmark
		&   \cmark
		& MLP \\
		
		
		CDFSL-V \cite{samarasinghe_cdfsl-v_2023}
		& cross-domain
		&  \xmark
		&   \xmark
		&  Logistic Regression \\
		
		
		\bottomrule
		
	\end{tabularx}
	
	\label{tab:model-comparison}
\end{table} 

\subsection{Datasets}
\Cref{sec:choice} shows and justifies the 25 different scenarios (base and novel combinations) used to evaluate each model. When a dataset is used within the base or novel set, the training and testing class splits are used, respectively. The validation class split of the base set is used for hyperparameter tuning. When the base set consists of more than one dataset, their respective splits are combined. When base and novel sets have shared action classes, the class is removed from the base set so as to not violate the few-shot assumption.

\subsection{Implementation Details}\label{sec:implementation-details}

\paragraph{Training and Evaluation.} To train STRM, MoLo, ProtoNet, and \textit{Transfer} models, a number of model parameters and training configurations are kept consistent for a fair comparison, whereby standard practices are employed. For CDFSL-V, we follow exactly the implementation and training details outlined in their paper \cite{samarasinghe_cdfsl-v_2023}. See \Cref{app:further-eval-details} for in-depth details on the training configuration for each model. For simplicity, we tune hyperparameters using the training and validation splits of the base dataset(s). This follows the conventional single-domain scenario. We conduct the few-shot evaluation on 10,000 of the same randomly sampled episodes (for consistency) and report the mean accuracy. We additionally evaluate an untrained, random weight initialised version of each model. Results were collected on a NVIDIA GeForce RTX 3090 Ti.

\paragraph{Episode Construction.} To train episodic models (STRM, MoLo, and ProtoNet) on \textit{N}-way \textit{K}-shot tasks, we construct the support set by first selecting (uniformly at random) which dataset to sample from within $\mathcal{D}_{base}$ (if $\mathcal{D}_{base}$ consists of multiple datasets), then select \textit{N} classes from within that dataset, then select \textit{K} examples of each class. When $\mathcal{D}_{base}$ consists of one dataset, this is the conventional method, and when $\mathcal{D}_{base}$ consists of multiple datasets, this ensures episodes correspond to realistic classification problems, as is performed in \cite{triantafillou_meta-dataset_2020}. To construct the query set we select \textit{N} samples, where each sample belongs to one of the \textit{N} classes in the support set. When evaluating models, $\mathcal{D}_{novel}$ is sampled from, following the same procedure as above.

\section{Experimental Results and Discussion}\label{sec:results}

Here, we present and analyse the 5-way 5-shot performance of each model under each of the 25 chosen evaluation settings. First, we compare and discuss each model's few-shot performance, then analyse the implications of learning from more data. We compare the overall few-shot performance with respect to the pre-calculated domain difference measured last. Refer to \Cref{app:1-shot-results} to see 5-way 1-shot results, which follow similar trends to those discussed below.

\subsection{Model Performance}

\begin{figure}[t]
	\centering
	\includegraphics[width=\linewidth]{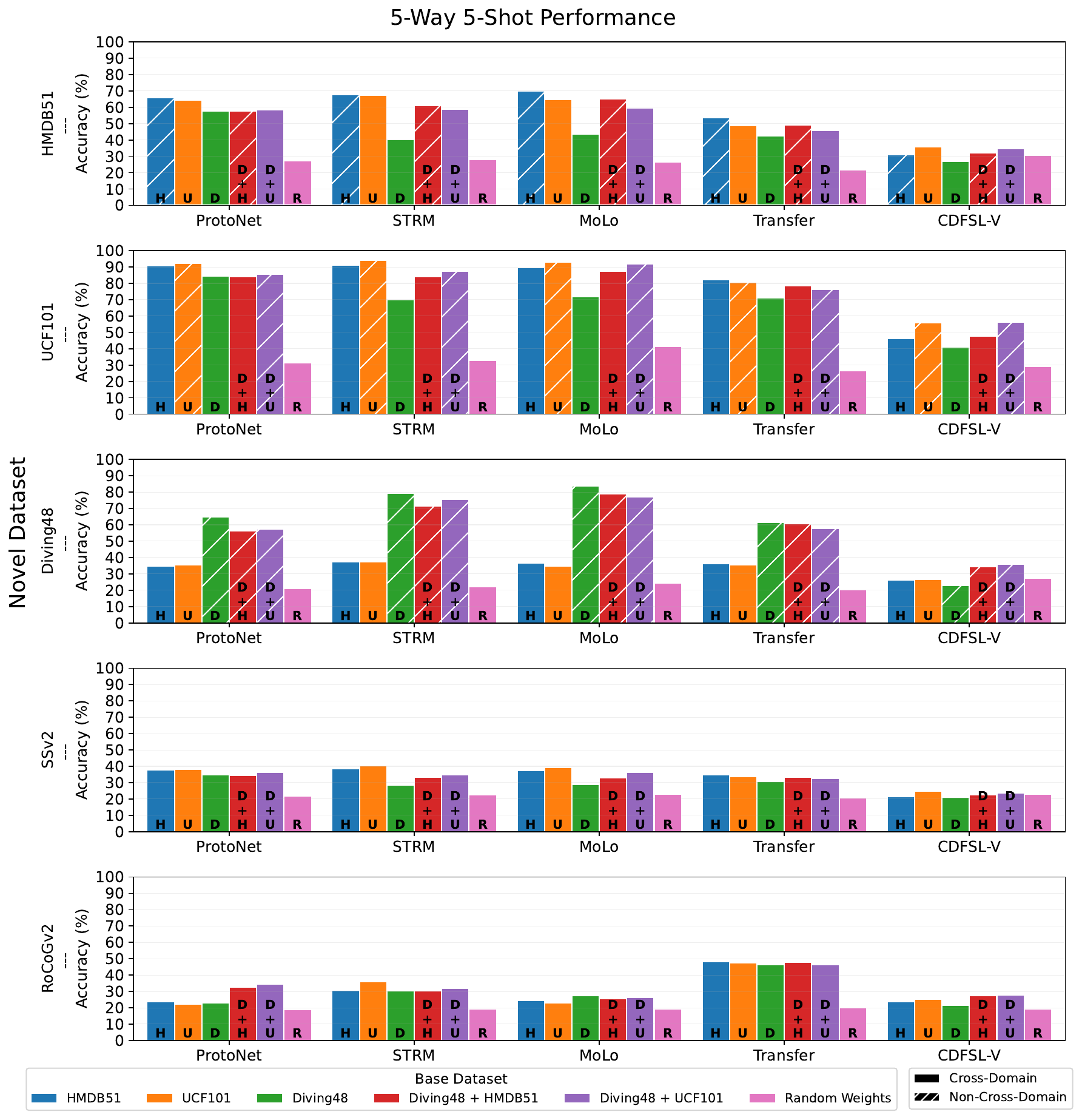}
	\caption{5-Way 5-Shot performance of existing single-domain (ProtoNet, STRM, MoLo), transfer-based (\textit{Transfer}), and cross-domain (CDFSL-V) FSAR models trained using different base datasets (indicated by colour), and evaluated on different novel datasets (each subfigure). Each model was also tested untrained with random weights (pink bars). The raw values are found in \Cref{app:raw_5shot_results}.}
	\label{fig:model-fs-performance}
\end{figure}

We compare the performances of each model to comment on which model aspects are effective for CD-FSAR and which need further development. \Cref{fig:model-fs-performance} presents a bar chart which displays the 5-way 5-shot performances achieved by each model (cluster of bars), for each of the 25 base dataset (colour of bar) and novel dataset (subfigure) combinations.

From the performance results on RoCoGv2, which has the largest domain difference to the other datasets, the simple transfer-learning based method considerably outperforms the single-domain and cross-domain methods by over 12 percentage points.  This reveals that as the domain difference increases, an adaptation stage where knowledge is transferred becomes increasingly important, and simple transfer learning methods remain effective even when only a few unseen examples are available. This finding is consistent with those deduced by Guo \etal \cite{vedaldi_broader_2020} and Chen \etal \cite{chen_closer_2019} for cross-domain few-shot image classification. The poor performance of single-domain methods in these cross-domain scenarios could be explained by their use of closest prototype classification in feature space and lack of adaptation. They assume that features discriminative of the base classes are also discriminative of novel classes, and thus they are evidently unable to generalise to out-of-distribution novel classes. This further highlights the importance of adapting learnt knowledge to an unseen domain.

Under these new, more challenging, cross-domain settings, the specialised cross-domain method (CDFSL-V) achieves the lowest performance, comparable to the random weight model performances in some instances. These results contradict the model creators original findings. In comparison to their original evaluation procedures which utilised only Kinetics-400 \cite{kinetics} (400 classes) as the base dataset, the datasets we use are more than 4x smaller. This indicates this method requires much more data to learn from to perform well. Additionally, while the existing CD-FSAR models see a substantial amount of unlabelled novel data during training, the poor performance suggests current self-supervised learning techniques for video data are inappropriate in this application. This could be attributed to the smaller size of the novel dataset, especially for RoCoGv2, which highlights the issue of assuming a substantial amount of unlabelled novel data is available during training. We note that Guo \etal \cite{vedaldi_broader_2020} chose to not compare to cross-domain methods which unfairly require unlabelled novel data. No specific CD-FSAR model exists which does not require unlabelled novel data. 

Comparing the performance of single-domain methods (ProtoNet, STRM, and MoLo), we witness ProtoNet performing comparably to, and in some cases better than, STRM and MoLo. Noting that ProtoNet does not measure temporal alignment between frames/frame subsequences while STRM and MoLo do, this finding suggests that existing SOTA temporal alignment techniques fail to generalise on unseen domains. This indicates that future endeavours should concentrate not only on how to transfer spatial information to new domains, but also temporal information.

\subsection{How does Base Dataset Composition Impact Few-Shot Performance?}

We investigate how base dataset composition can impact downstream few-shot performance, in terms of the use of multiple-datasets and use of data from the same domain as the novel dataset. In \Cref{fig:model-fs-performance}, cross-hatching on bars indicates a combination where data from the same domain as the novel set is included in the base set, and is thus not a cross-domain setting.

Under cross-domain settings (i.e., solid bars), when more data is added to the base dataset (i.e., the performance on D vs. H vs. D+H), in some instances, such as ProtoNet tested on RoCoGv2, we witness a performance increase, in other instances, such as STRM tested on RoCoGv2, we observe similar performance, and in most other instances, such as MoLo tested on SSv2, we find the multi-domain performance tends to be an average of the performance on the individual base sets. This reveals issues with sensitivity to base class construction, where as we learn from more data with more diversity, performance does not necessarily improve. This coincides with the findings of Luo \etal \cite{luo_closer_2023} for image classification, who found that larger base datasets may lead to degraded performance on certain downstream datasets. This insight contradicts the common finding in machine learning, where learning from more data leads to better generalisation. We hope this initial finding prompts research towards gaining a better understanding of how the base dataset construction influences downstream FSAR performance, as is studied in detail for few-shot image classification \cite{vedaldi_impact_2020, luo_closer_2023}.

We also observe that when the base and novel sets contain samples from the same domain (i.e., hatched bars), performance is boosted, especially for single-domain models (ProtoNet, STRM, and MoLo). This demonstrates the bias introduced when learning from data sampled from the same domain used for model evaluation . 

\begin{figure}[t]
	\includegraphics[width=\textwidth]{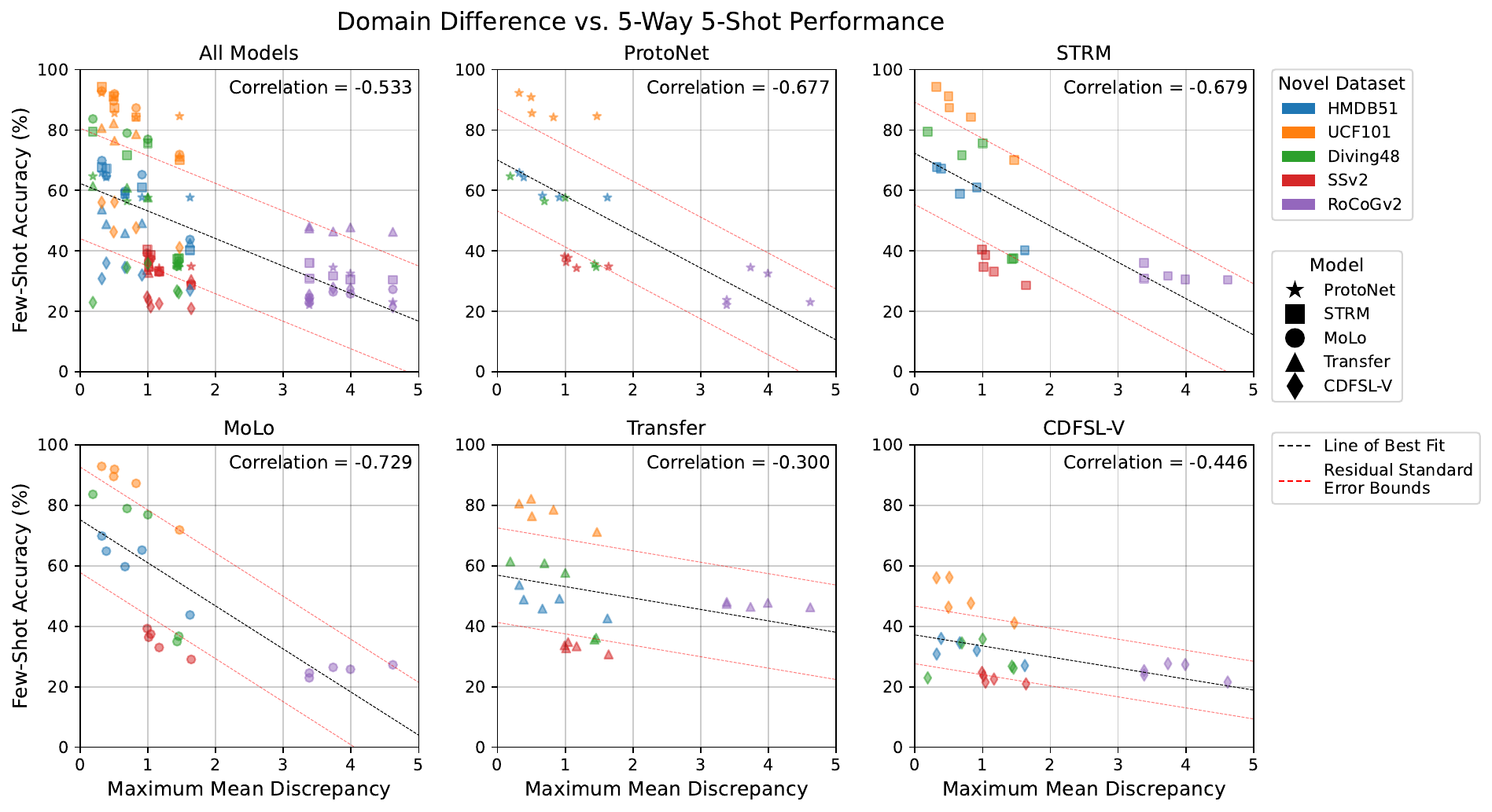}
	\caption{Measure of domain difference against downstream few-shot performance for each model.}
	\label{fig:domain-diff-vs-fewshot-performance}
\end{figure}

\subsection{How does Domain Difference Influence Few-Shot Performance?} 

We analyse the relationship between the pre-calculated domain difference measures (MMD) and downstream few-shot performance of each model under each scenario. \Cref{fig:domain-diff-vs-fewshot-performance} plots this relationship for each model separately and all models together (top left), and displays each measured correlation.

As indicated, we witness a negative correlation between few-shot performance and domain difference for the single-domain methods (ProtoNet, STRM, and MoLo), where as the domain difference increases, the downstream few-shot performance decreases. The transfer-based and cross-domain approaches also follow this trend, but with a weaker correlation. This finding is similar to that of Sbai \etal \cite{vedaldi_impact_2020}, who deduced that the similarity of the base classes and novel classes has a crucial effect on few-shot performance, however their finding was simply based on qualitative similarity. This result indicates this simple measure could be used to predict downstream performance and hence inform which dataset or combinations of datasets to select as the base dataset for a given a novel dataset. In this way, the combination with the smallest domain difference to the novel dataset should be chosen. 

In \Cref{fig:domain-diff-vs-fewshot-performance} we plot the line of best fit and residual standard error bounds. As seen here, some trends are not linear, and hence more analysis is required to inform if there is a better predictive function. We also observe that model performance on SSv2 (red) is comparably low, even for the transfer-learning approach. This suggests that the distribution of the novel dataset also impacts few-shot performance. 

In this preliminary investigation we use MMD since it is commonly used in domain adaptation works and can provide some initial insights. However, this measure is limited in its ability to summarise non-spatial variations, which are important for video datasets. Thus, future work should evaluate whether different measures which capture the temporal differences between domains are better indicators of downstream few-shot performance.

\section{Conclusion}

In this paper, we conducted a systematic empirical study to understand the cross-domain capabilities of a diverse set of existing single-domain, transfer-based, and cross-domain FSAR models. Each model was evaluated on new cross-domain scenarios which were selected based on their domain difference (MMD). Our research uncovers useful insights and challenges in the field of CD-FSAR (summarised below), and we hope to inspire future work in these directions, to help move FSAR toward real-world usability.

Our results demonstrated that as the domain difference increases, the simple transfer-learning based method outperforms the single-domain and cross-domain methods by over 12 percentage points. Under these more challenging settings, the specialised cross-domain method achieves the lowest performance, comparable to random weight model performances, and does not perform well when the novel dataset is relatively small. This highlights that future CD-FSAR model development should not assume a substantial amount of unlabelled novel data is available at training time. We also find the SOTA single-domain FSAR models which use temporal alignment achieve similar or worse performance than earlier methods which omit temporal alignment, suggesting existing temporal alignment techniques fail to generalise on unseen domains. We also witness a correlation between domain difference and downstream performance, suggesting this simple measure could be used to inform base dataset design. However, the correlation is not perfect, and thus the measure could be improved. Lastly, in some instances, adding more datasets to the base set boosts few-shot performance, whereas in others it reduces, revealing a challenging sensitivity to base class construction.

\section*{Acknowledgement}
This work was supported by the Australian Centre for Robotics, the Rio Tinto Centre for Mine Automation, and the Australian Government Research Training Program.


\def\url#1{}
\bibliographystyle{IEEEtran}
\bibliography{bib-cross-domain, bib-Datasets, bib-models, bib-Applications, bib-peripheral}
\newpage


\appendix

\section{Example Samples from Each Dataset}\label{app:dataset_examples}

\begin{figure}[H]
	\centering
	
	\begin{subfigure}[t]{\linewidth}
			\begin{subfigure}[t]{.16\linewidth}
					\centering\includegraphics[width=.95\linewidth]{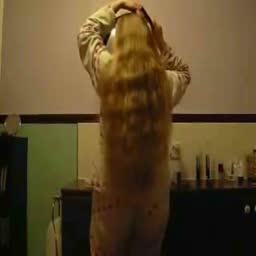}
			\end{subfigure}
			\begin{subfigure}[t]{.16\linewidth}
				\centering\includegraphics[width=.95\linewidth]{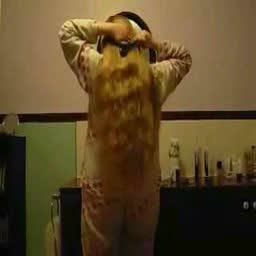}
			\end{subfigure}
			\begin{subfigure}[t]{.16\linewidth}
				\centering\includegraphics[width=.95\linewidth]{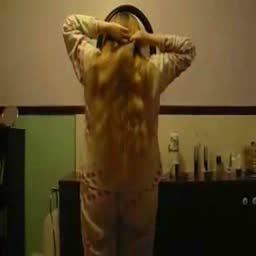}
			\end{subfigure}
			\begin{subfigure}[t]{.16\linewidth}
				\centering\includegraphics[width=.95\linewidth]{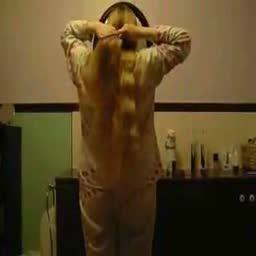}
			\end{subfigure}
			\begin{subfigure}[t]{.16\linewidth}
				\centering\includegraphics[width=.95\linewidth]{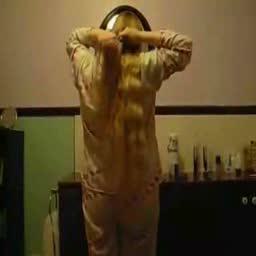}
			\end{subfigure}
			\begin{subfigure}[t]{.16\linewidth}
				\centering\includegraphics[width=.95\linewidth]{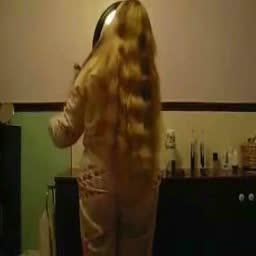}
			\end{subfigure}
			\caption{Brush Hair}
	\end{subfigure}
	
	\medskip
	
	\begin{subfigure}[t]{\linewidth}
		\begin{subfigure}[t]{.16\linewidth}
			\centering\includegraphics[width=.95\linewidth]{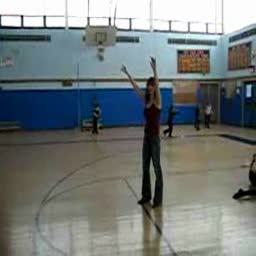}
		\end{subfigure}
		\begin{subfigure}[t]{.16\linewidth}
			\centering\includegraphics[width=.95\linewidth]{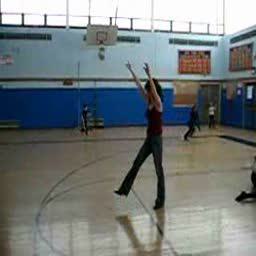}
		\end{subfigure}
		\begin{subfigure}[t]{.16\linewidth}
			\centering\includegraphics[width=.95\linewidth]{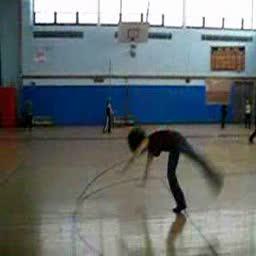}
		\end{subfigure}
		\begin{subfigure}[t]{.16\linewidth}
			\centering\includegraphics[width=.95\linewidth]{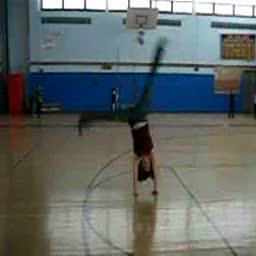}
		\end{subfigure}
		\begin{subfigure}[t]{.16\linewidth}
			\centering\includegraphics[width=.95\linewidth]{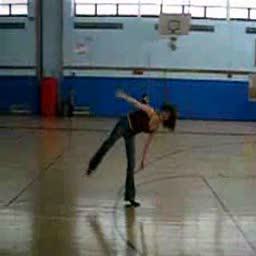}
		\end{subfigure}
		\begin{subfigure}[t]{.16\linewidth}
			\centering\includegraphics[width=.95\linewidth]{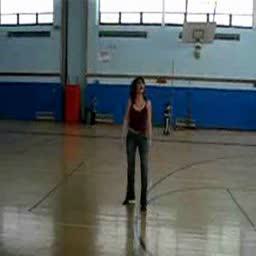}
		\end{subfigure}
		\caption{Cartwheel}
	\end{subfigure}
	
	\medskip
	
	\begin{subfigure}[t]{\linewidth}
		\begin{subfigure}[t]{.16\linewidth}
			\centering\includegraphics[width=.95\linewidth]{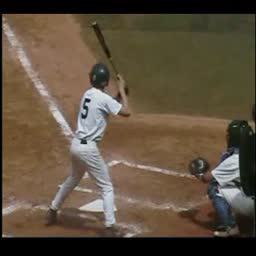}
		\end{subfigure}
		\begin{subfigure}[t]{.16\linewidth}
			\centering\includegraphics[width=.95\linewidth]{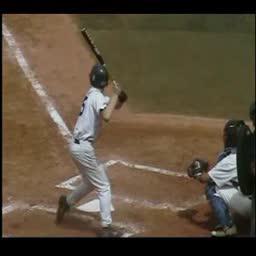}
		\end{subfigure}
		\begin{subfigure}[t]{.16\linewidth}
			\centering\includegraphics[width=.95\linewidth]{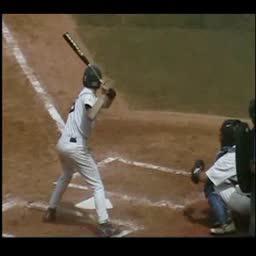}
		\end{subfigure}
		\begin{subfigure}[t]{.16\linewidth}
			\centering\includegraphics[width=.95\linewidth]{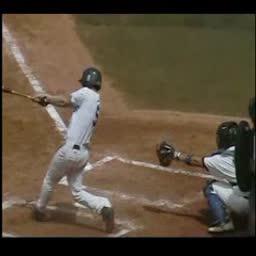}
		\end{subfigure}
		\begin{subfigure}[t]{.16\linewidth}
			\centering\includegraphics[width=.95\linewidth]{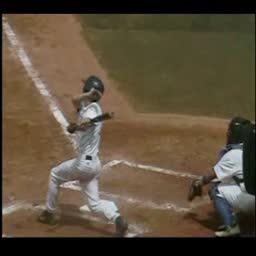}
		\end{subfigure}
		\begin{subfigure}[t]{.16\linewidth}
			\centering\includegraphics[width=.95\linewidth]{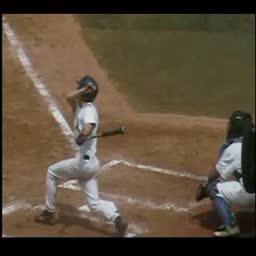}
		\end{subfigure}
		\caption{Swing Baseball}
	\end{subfigure}
	
	\caption{Example frame sequences from videos within the HMDB51 dataset, with their corresponding action label.}
	\label{fig:HMDB51}
\end{figure}

\begin{figure}[H]
	\centering
	
	\begin{subfigure}[t]{\linewidth}
			\begin{subfigure}[t]{.16\linewidth}
					\centering\includegraphics[width=.95\linewidth]{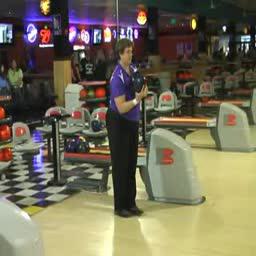}
			\end{subfigure}
			\begin{subfigure}[t]{.16\linewidth}
				\centering\includegraphics[width=.95\linewidth]{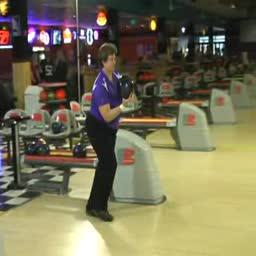}
			\end{subfigure}
			\begin{subfigure}[t]{.16\linewidth}
				\centering\includegraphics[width=.95\linewidth]{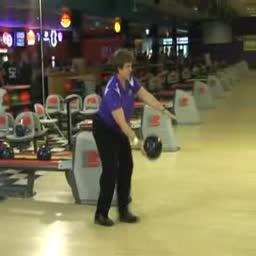}
			\end{subfigure}
			\begin{subfigure}[t]{.16\linewidth}
				\centering\includegraphics[width=.95\linewidth]{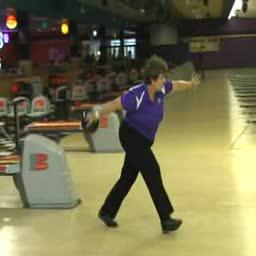}
			\end{subfigure}
			\begin{subfigure}[t]{.16\linewidth}
				\centering\includegraphics[width=.95\linewidth]{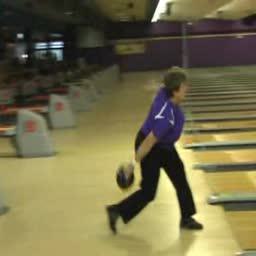}
			\end{subfigure}
			\begin{subfigure}[t]{.16\linewidth}
				\centering\includegraphics[width=.95\linewidth]{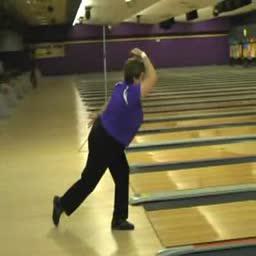}
			\end{subfigure}
			\caption{Bowling}
	\end{subfigure}
	
	\medskip
	
	\begin{subfigure}[t]{\linewidth}
		\begin{subfigure}[t]{.16\linewidth}
			\centering\includegraphics[width=.95\linewidth]{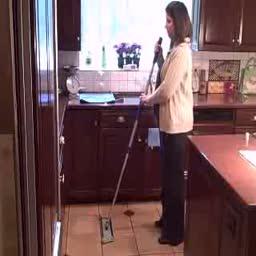}
		\end{subfigure}
		\begin{subfigure}[t]{.16\linewidth}
			\centering\includegraphics[width=.95\linewidth]{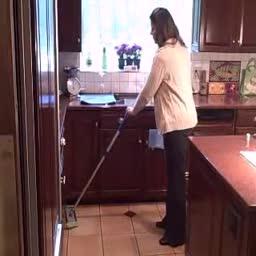}
		\end{subfigure}
		\begin{subfigure}[t]{.16\linewidth}
			\centering\includegraphics[width=.95\linewidth]{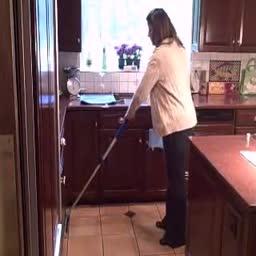}
		\end{subfigure}
		\begin{subfigure}[t]{.16\linewidth}
			\centering\includegraphics[width=.95\linewidth]{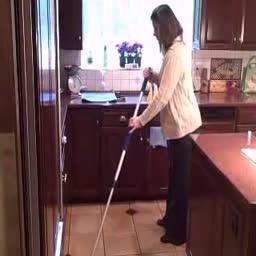}
		\end{subfigure}
		\begin{subfigure}[t]{.16\linewidth}
			\centering\includegraphics[width=.95\linewidth]{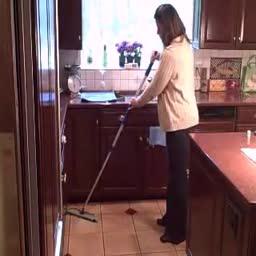}
		\end{subfigure}
		\begin{subfigure}[t]{.16\linewidth}
			\centering\includegraphics[width=.95\linewidth]{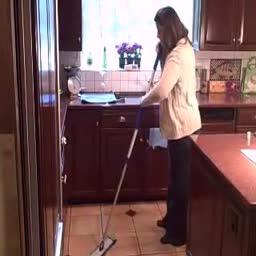}
		\end{subfigure}
		\caption{Mopping Floor}
	\end{subfigure}
	
	\medskip
	
	\begin{subfigure}[t]{\linewidth}
		\begin{subfigure}[t]{.16\linewidth}
			\centering\includegraphics[width=.95\linewidth]{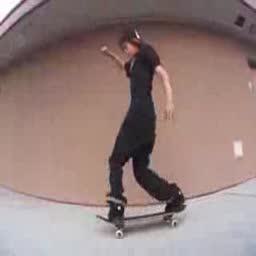}
		\end{subfigure}
		\begin{subfigure}[t]{.16\linewidth}
			\centering\includegraphics[width=.95\linewidth]{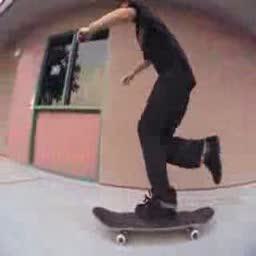}
		\end{subfigure}
		\begin{subfigure}[t]{.16\linewidth}
			\centering\includegraphics[width=.95\linewidth]{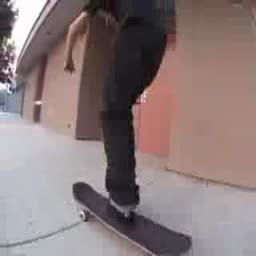}
		\end{subfigure}
		\begin{subfigure}[t]{.16\linewidth}
			\centering\includegraphics[width=.95\linewidth]{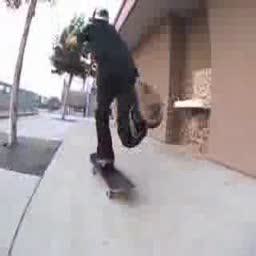}
		\end{subfigure}
		\begin{subfigure}[t]{.16\linewidth}
			\centering\includegraphics[width=.95\linewidth]{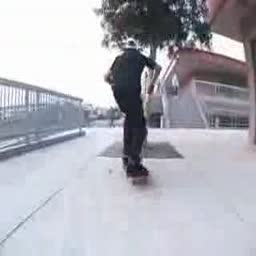}
		\end{subfigure}
		\begin{subfigure}[t]{.16\linewidth}
			\centering\includegraphics[width=.95\linewidth]{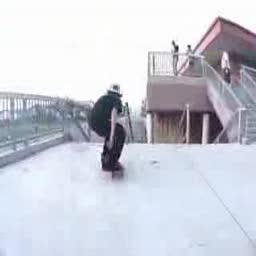}
		\end{subfigure}
		\caption{Skateboarding}
	\end{subfigure}
	
	\caption{Example frame sequences from videos within the UCF101 dataset, with their corresponding action label.}
	
	\label{fig:UCF101}
\end{figure}

\begin{figure}[H]
	\centering
	
	\begin{subfigure}[t]{\linewidth}
			\begin{subfigure}[t]{.16\linewidth}
					\centering\includegraphics[width=.95\linewidth]{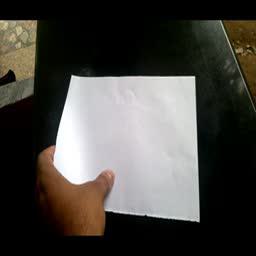}
			\end{subfigure}
			\begin{subfigure}[t]{.16\linewidth}
				\centering\includegraphics[width=.95\linewidth]{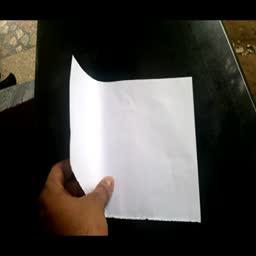}
			\end{subfigure}
			\begin{subfigure}[t]{.16\linewidth}
				\centering\includegraphics[width=.95\linewidth]{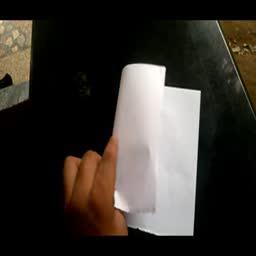}
			\end{subfigure}
			\begin{subfigure}[t]{.16\linewidth}
				\centering\includegraphics[width=.95\linewidth]{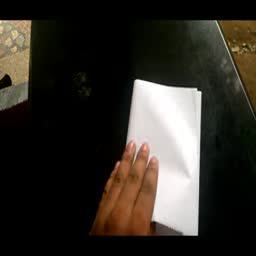}
			\end{subfigure}
			\begin{subfigure}[t]{.16\linewidth}
				\centering\includegraphics[width=.95\linewidth]{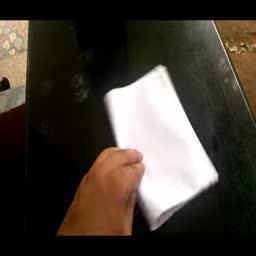}
			\end{subfigure}
			\begin{subfigure}[t]{.16\linewidth}
				\centering\includegraphics[width=.95\linewidth]{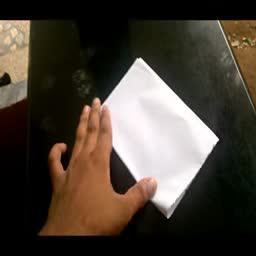}
			\end{subfigure}
			\caption{Folding \textit{something}}
	\end{subfigure}
	
	\medskip
	
	\begin{subfigure}[t]{\linewidth}
		\begin{subfigure}[t]{.16\linewidth}
			\centering\includegraphics[width=.95\linewidth]{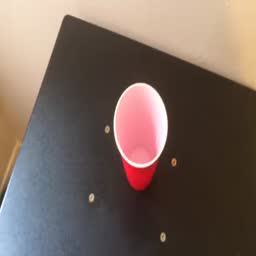}
		\end{subfigure}
		\begin{subfigure}[t]{.16\linewidth}
			\centering\includegraphics[width=.95\linewidth]{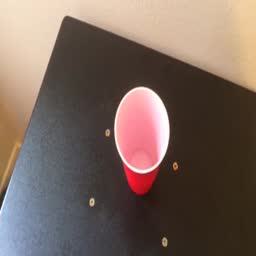}
		\end{subfigure}
		\begin{subfigure}[t]{.16\linewidth}
			\centering\includegraphics[width=.95\linewidth]{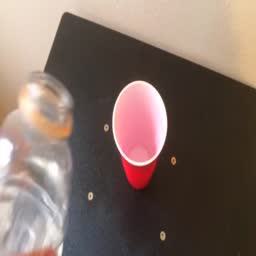}
		\end{subfigure}
		\begin{subfigure}[t]{.16\linewidth}
			\centering\includegraphics[width=.95\linewidth]{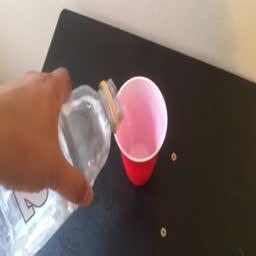}
		\end{subfigure}
		\begin{subfigure}[t]{.16\linewidth}
			\centering\includegraphics[width=.95\linewidth]{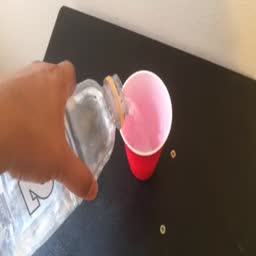}
		\end{subfigure}
		\begin{subfigure}[t]{.16\linewidth}
			\centering\includegraphics[width=.95\linewidth]{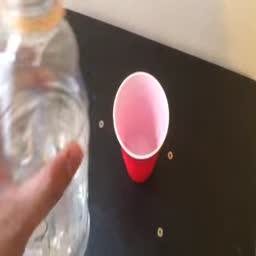}
		\end{subfigure}
		\caption{Pouring \textit{something} into \textit{something}}
	\end{subfigure}
	
	\medskip
	
	\begin{subfigure}[t]{\linewidth}
		\begin{subfigure}[t]{.16\linewidth}
			\centering\includegraphics[width=.95\linewidth]{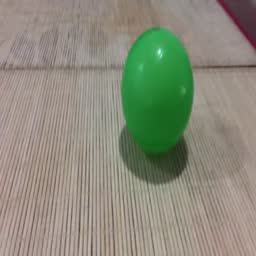}
		\end{subfigure}
		\begin{subfigure}[t]{.16\linewidth}
			\centering\includegraphics[width=.95\linewidth]{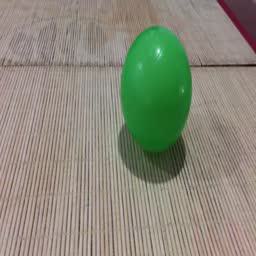}
		\end{subfigure}
		\begin{subfigure}[t]{.16\linewidth}
			\centering\includegraphics[width=.95\linewidth]{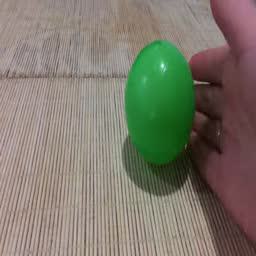}
		\end{subfigure}
		\begin{subfigure}[t]{.16\linewidth}
			\centering\includegraphics[width=.95\linewidth]{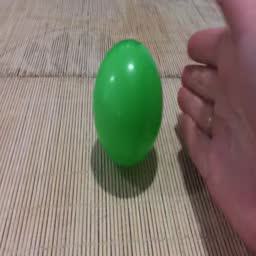}
		\end{subfigure}
		\begin{subfigure}[t]{.16\linewidth}
			\centering\includegraphics[width=.95\linewidth]{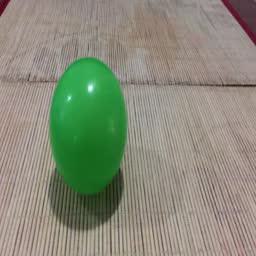}
		\end{subfigure}
		\begin{subfigure}[t]{.16\linewidth}
			\centering\includegraphics[width=.95\linewidth]{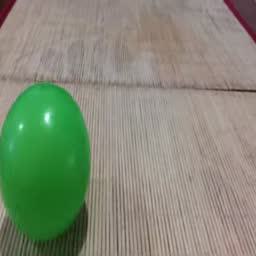}
		\end{subfigure}
		\caption{Pushing \textit{something} from right to left}
	\end{subfigure}
	
	\caption{Example frame sequences from videos within the SSv2 dataset, with their corresponding action label.}
	\label{fig:SSv2}
\end{figure}

\begin{figure}[H]
	\centering
	
	\begin{subfigure}[t]{\linewidth}
		\begin{subfigure}[t]{.16\linewidth}
			\centering\includegraphics[width=.95\linewidth]{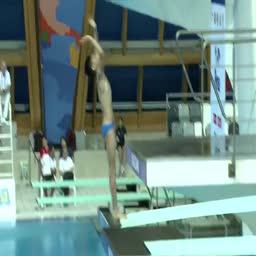}
		\end{subfigure}
		\begin{subfigure}[t]{.16\linewidth}
			\centering\includegraphics[width=.95\linewidth]{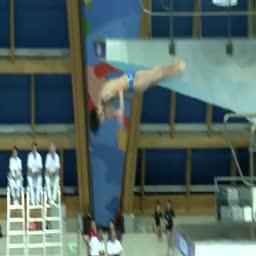}
		\end{subfigure}
		\begin{subfigure}[t]{.16\linewidth}
			\centering\includegraphics[width=.95\linewidth]{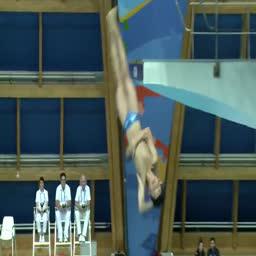}
		\end{subfigure}
		\begin{subfigure}[t]{.16\linewidth}
			\centering\includegraphics[width=.95\linewidth]{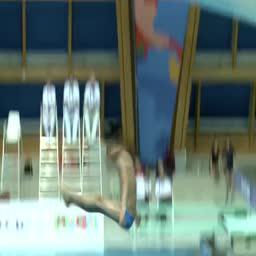}
		\end{subfigure}
		\begin{subfigure}[t]{.16\linewidth}
			\centering\includegraphics[width=.95\linewidth]{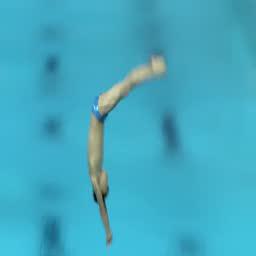}
		\end{subfigure}
		\begin{subfigure}[t]{.16\linewidth}
			\centering\includegraphics[width=.95\linewidth]{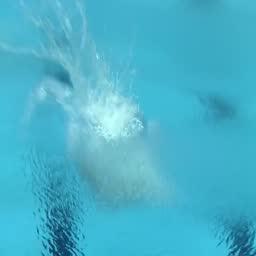}
		\end{subfigure}
		\caption{Back\_15som\_25Twis\_FREE}
	\end{subfigure}
	
	\medskip
	
	\begin{subfigure}[t]{\linewidth}
		\begin{subfigure}[t]{.16\linewidth}
			\centering\includegraphics[width=.95\linewidth]{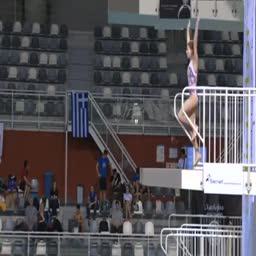}
		\end{subfigure}
		\begin{subfigure}[t]{.16\linewidth}
			\centering\includegraphics[width=.95\linewidth]{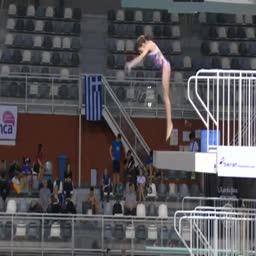}
		\end{subfigure}
		\begin{subfigure}[t]{.16\linewidth}
			\centering\includegraphics[width=.95\linewidth]{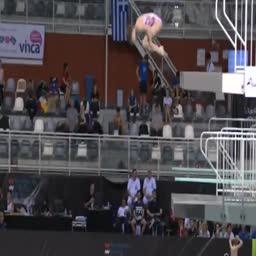}
		\end{subfigure}
		\begin{subfigure}[t]{.16\linewidth}
			\centering\includegraphics[width=.95\linewidth]{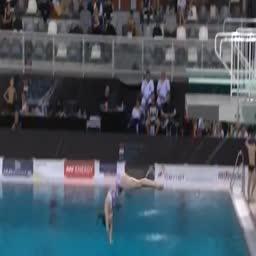}
		\end{subfigure}
		\begin{subfigure}[t]{.16\linewidth}
			\centering\includegraphics[width=.95\linewidth]{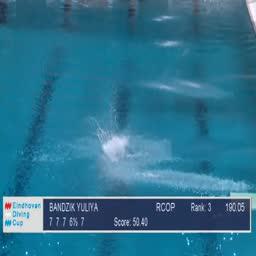}
		\end{subfigure}
		\begin{subfigure}[t]{.16\linewidth}
			\centering\includegraphics[width=.95\linewidth]{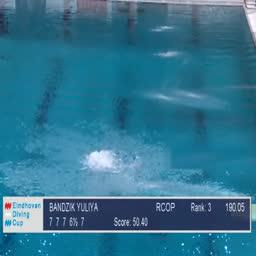}
		\end{subfigure}
		\caption{Forward\_25som\_NoTwis\_TUCK}
	\end{subfigure}
	
	\medskip
	
	\begin{subfigure}[t]{\linewidth}
		\begin{subfigure}[t]{.16\linewidth}
			\centering\includegraphics[width=.95\linewidth]{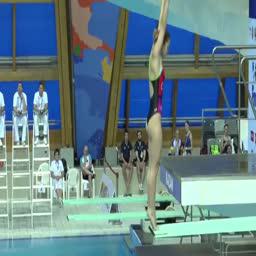}
		\end{subfigure}
		\begin{subfigure}[t]{.16\linewidth}
			\centering\includegraphics[width=.95\linewidth]{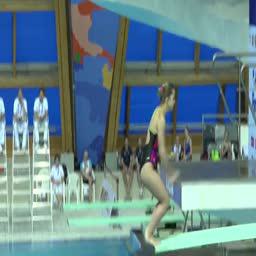}
		\end{subfigure}
		\begin{subfigure}[t]{.16\linewidth}
			\centering\includegraphics[width=.95\linewidth]{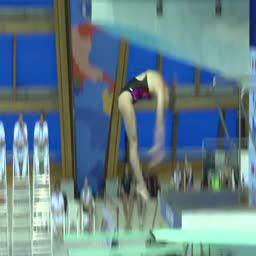}
		\end{subfigure}
		\begin{subfigure}[t]{.16\linewidth}
			\centering\includegraphics[width=.95\linewidth]{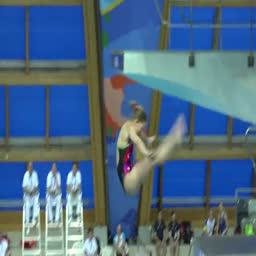}
		\end{subfigure}
		\begin{subfigure}[t]{.16\linewidth}
			\centering\includegraphics[width=.95\linewidth]{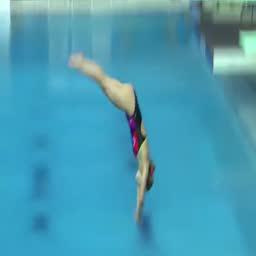}
		\end{subfigure}
		\begin{subfigure}[t]{.16\linewidth}
			\centering\includegraphics[width=.95\linewidth]{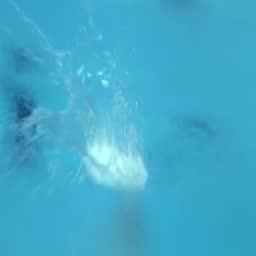}
		\end{subfigure}
		\caption{Inward\_15som\_NoTwis\_PIKE}
	\end{subfigure}
	
	\caption{Example frame sequences from videos within the Diving48 dataset, with their corresponding action label.}
	\label{fig:Diving48}
\end{figure}

\begin{figure}[H]
	\centering
	
	\begin{subfigure}[t]{\linewidth}
		\begin{subfigure}[t]{.16\linewidth}
			\centering\includegraphics[width=.95\linewidth]{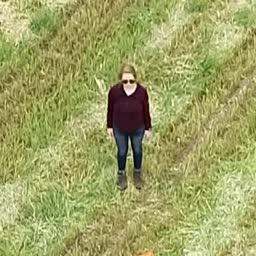}
		\end{subfigure}
		\begin{subfigure}[t]{.16\linewidth}
			\centering\includegraphics[width=.95\linewidth]{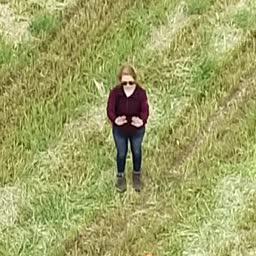}
		\end{subfigure}
		\begin{subfigure}[t]{.16\linewidth}
			\centering\includegraphics[width=.95\linewidth]{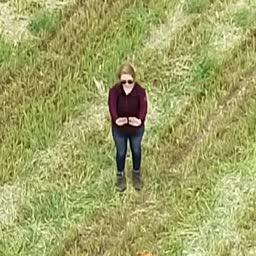}
		\end{subfigure}
		\begin{subfigure}[t]{.16\linewidth}
			\centering\includegraphics[width=.95\linewidth]{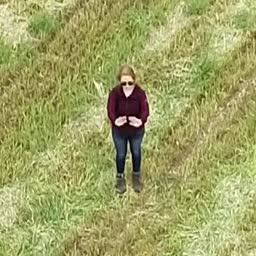}
		\end{subfigure}
		\begin{subfigure}[t]{.16\linewidth}
			\centering\includegraphics[width=.95\linewidth]{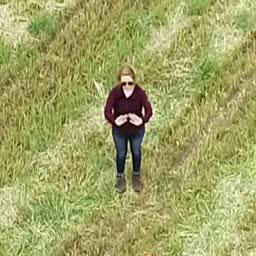}
		\end{subfigure}
		\begin{subfigure}[t]{.16\linewidth}
			\centering\includegraphics[width=.95\linewidth]{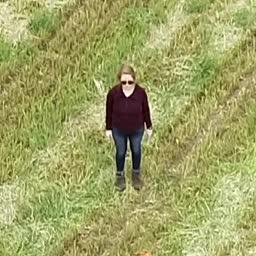}
		\end{subfigure}
		\caption{Move In Reverse}
	\end{subfigure}
	
	\medskip
	
	\begin{subfigure}[t]{\linewidth}
			\begin{subfigure}[t]{.16\linewidth}
					\centering\includegraphics[width=.95\linewidth]{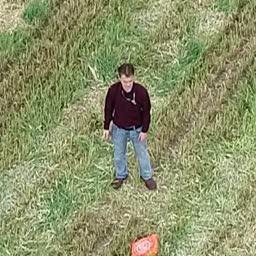}
				\end{subfigure}
			\begin{subfigure}[t]{.16\linewidth}
				\centering\includegraphics[width=.95\linewidth]{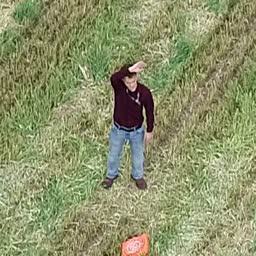}
			\end{subfigure}
			\begin{subfigure}[t]{.16\linewidth}
				\centering\includegraphics[width=.95\linewidth]{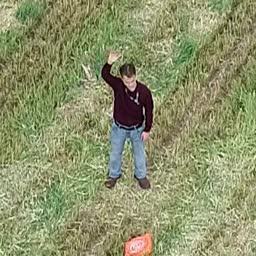}
			\end{subfigure}
			\begin{subfigure}[t]{.16\linewidth}
				\centering\includegraphics[width=.95\linewidth]{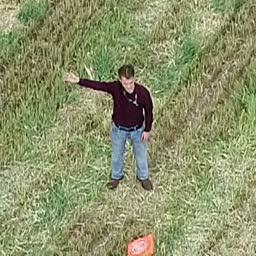}
			\end{subfigure}
			\begin{subfigure}[t]{.16\linewidth}
				\centering\includegraphics[width=.95\linewidth]{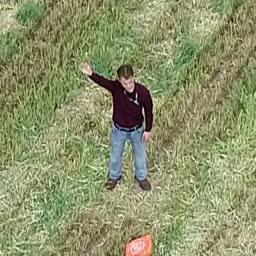}
			\end{subfigure}
			\begin{subfigure}[t]{.16\linewidth}
				\centering\includegraphics[width=.95\linewidth]{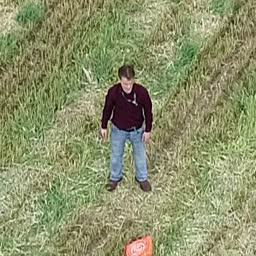}
			\end{subfigure}
			\caption{Attention}
		\end{subfigure}
	
	\medskip
	
		\begin{subfigure}[t]{\linewidth}
				\begin{subfigure}[t]{.16\linewidth}
						\centering\includegraphics[width=.95\linewidth]{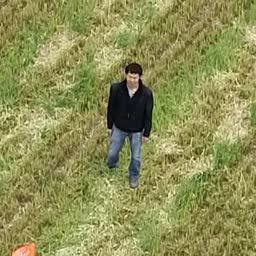}
				\end{subfigure}
				\begin{subfigure}[t]{.16\linewidth}
				\centering\includegraphics[width=.95\linewidth]{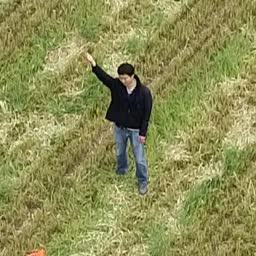}
				\end{subfigure}
				\begin{subfigure}[t]{.16\linewidth}
				\centering\includegraphics[width=.95\linewidth]{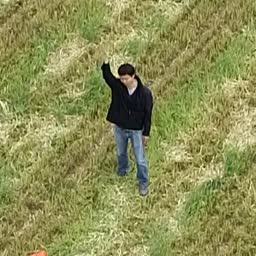}
				\end{subfigure}
				\begin{subfigure}[t]{.16\linewidth}
				\centering\includegraphics[width=.95\linewidth]{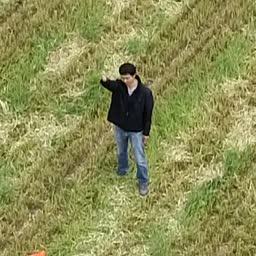}
				\end{subfigure}
				\begin{subfigure}[t]{.16\linewidth}
				\centering\includegraphics[width=.95\linewidth]{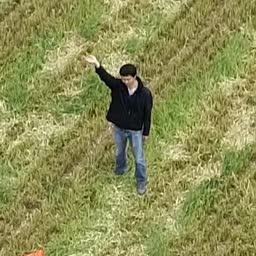}
				\end{subfigure}
				\begin{subfigure}[t]{.16\linewidth}
				\centering\includegraphics[width=.95\linewidth]{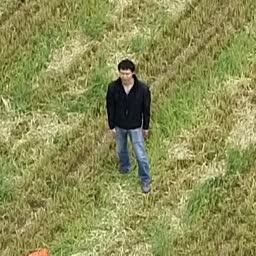}
				\end{subfigure}
				\caption{Rally}
			\end{subfigure}
	
	\caption{Example frame sequences from videos within the RoCoGv2 dataset, with their corresponding action label.}
	\label{fig:RoCoGv2}
\end{figure}

Figures \ref{fig:HMDB51} to \ref{fig:RoCoGv2} provides examples of 3 videos randomly sampled from each of the 5 datasets used in this study, where 6 uniformly sampled frames within each  video are shown, alongside their class label. These examples illustrate each dataset's qualitative characteristics, where we endeavoured to analyse datasets which span a variety of human action concepts (full-body motion in everyday activities, full-body motion in sports, gestures, and human-object interactions).

\vspace{-0.2cm}

In \Cref{fig:HMDB51}, the Human-Motion Database (HMDB51) \cite{kuehne_hmdb_2011} consists of everyday human actions, often involving full-body motion, recorded from varying perspectives with different backgrounds since videos are collated from movies and the internet. This dataset has potential for background biases to be introduced, which help to inform the action taking place.

In \Cref{fig:UCF101}, UCF101 \cite{soomro_ucf101_2012} contains the most diverse action categories, which can involve body-motion only, human-object interaction, and sports, as shown here. This dataset was collated from YouTube, and hence also has videos recorded with different perspectives, camera movements, and backgrounds. Similar to HMDB51, background biases which help to inform the action taking place exist for some action categories.

In \Cref{fig:SSv2}, the Something Something video database \cite{goyal_something_2017} contains action categories involving human-object interaction which take place in everyday tasks. As seen in these examples, classes typically involve doing `something' to `something', and thus require temporal reasoning. Often the video perspective is ego-centric, containing hands only, however some videos do contain humans, also performing hand motions with objects.

In \Cref{fig:Diving48}, Diving48 \cite{ferrari_resound_2018} consists of fine-grained actions unique to competitive diving. As seen, this dataset eliminates biases which help to inform the action class, since the background and object interactions simply consists of a board, pool, and spectators. As shown, the classes within this dataset are visually similar, and thus this dataset requires finer-grained temporal understanding due to the similarities between classes and lack of background bias. 

In \Cref{fig:RoCoGv2}, the Robot Control Gestures dataset \cite{reddy_synthetic--real_2023}, which consists of humans performing arm motion gestures intended for commanding a drone, also eliminates background biases since all videos contain a plain grass background and are recorded from an aerial perspective, as shown. These arm motion gestures are specific to the application. As seen, these gestures are similar and thus this dataset also requires finer-grained temporal understanding due to the similarities between classes and lack of background bias. 

\newpage
\section{Dataset Distributions}\label{app:dataset-stats}

Here, we provide more details on the distribution of classes and samples per class within each dataset. \Cref{tab:dataset-summary-statistics} provides a breakdown of the number of classes and samples within each of the training, validation, and test set splits, and overall, for each dataset. These splits are the most commonly used in literature, or created for this paper, as indicated. RoCoGv2 is only used as the novel set, and is thus not split. When a dataset is used within the base set, we utilised its training set during meta-training and its validation set when hyperparameter tuning. When a dataset is used as the novel set, we utilise its testing set split. In our study, combinations of HMDB51, UCF101, and Diving48 datasets were used within the base dataset, and all 5 datasets were used as individual novel sets. 

\begin{table}[H]
	\centering
	\renewcommand{\arraystretch}{1.3}
	\footnotesize
	\caption{Summary of the class and video composition of each dataset. }
	\vspace{0.2cm}
	\begin{tabularx}{\textwidth}{l *{5}{Y}}
		
		\Cline{2-6}{0.08em}
		
		
		& HMDB51
		& UCF101
		& SSv2{\small (-Full)}
		& Diving48 
		& RoCoGv2   \\
		
		
		\toprule
		
		
		Total Classes
		& 51
		&  101
		&   100
		&  47
		&  7  \\
		
		Total Video
		& 6766
		&  13320
		&   71796 
		&  16997
		&  178  \\
		
		\hdashline
		
		
		Training Classes
		& 31
		&  70
		&  64
		&  28
		&  -  \\
		
		Training Videos
		& 4280
		&  9154
		&  67013
		&  7326
		&  -  \\
		
		\hdashline
		
		
		Validation Classes
		& 10
		&  10
		&   12
		&  9
		&  -  \\
		
		Validation Videos
		& 1194
		&  1421
		&  1926
		&  4274
		&  -  \\
		
		\hdashline
		
		
		Testing Classes
		& 10
		&  21
		&   24
		&  10
		&  -  \\
		
		Testing Videos
		& 1292
		&  2745
		&  2857
		&  5497
		&  -  \\
		
		\hdashline
		
		
		Train/Val/Test Split Source
		& \cite{cao_few-shot_2020}
		&  \cite{cao_few-shot_2020}
		&  \cite{zhu_compound_2018}
		&  This paper
		&  -  \\
		
		
		\bottomrule
		
	\end{tabularx}
	
	\label{tab:dataset-summary-statistics}
	\vspace{-0.2cm}
\end{table} 

\begin{figure}[H]
	\centering
	\begin{subfigure}{0.32\linewidth}
		\includegraphics[width=\linewidth]{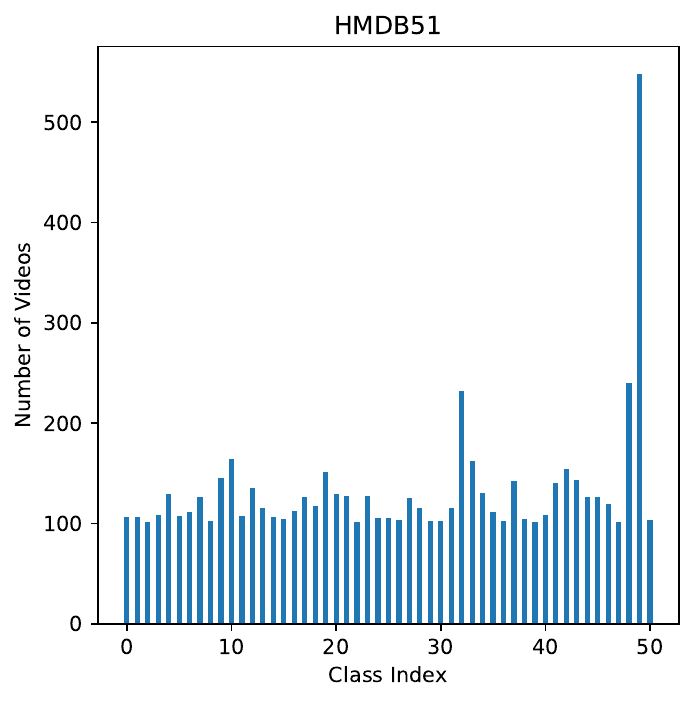}
	\end{subfigure}
	\begin{subfigure}{0.32\linewidth}
		\includegraphics[width=\linewidth]{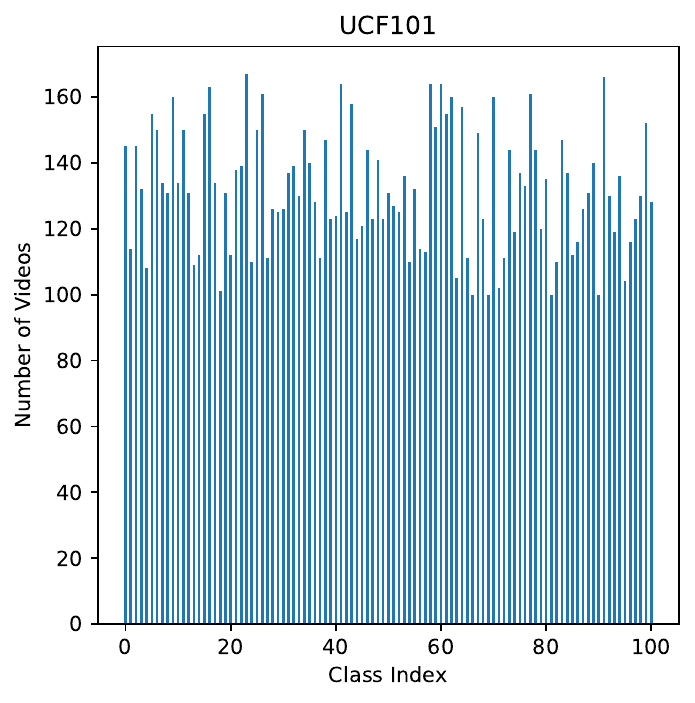}
	\end{subfigure}
	\begin{subfigure}{0.32\linewidth}
		\includegraphics[width=\linewidth]{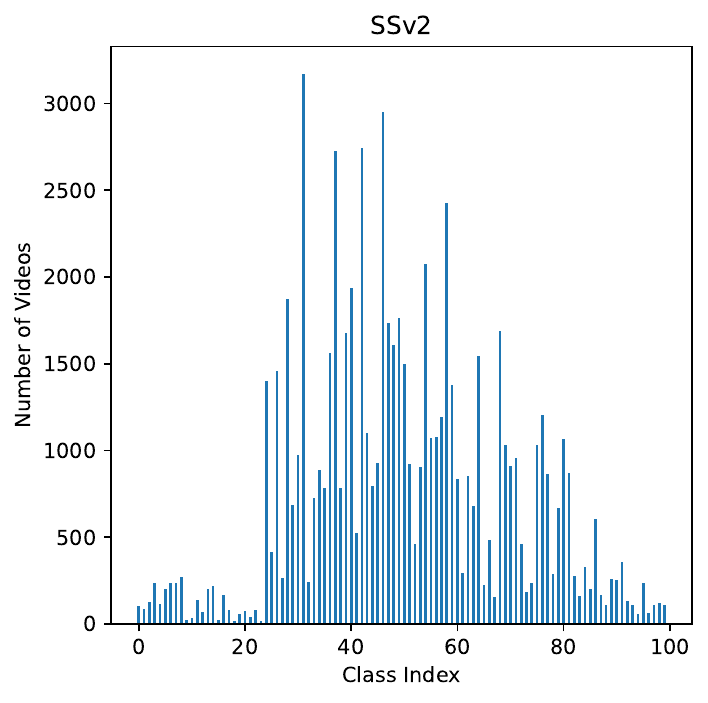}
	\end{subfigure}
	\vfill
	\begin{subfigure}{0.32\linewidth}
		\includegraphics[width=\linewidth]{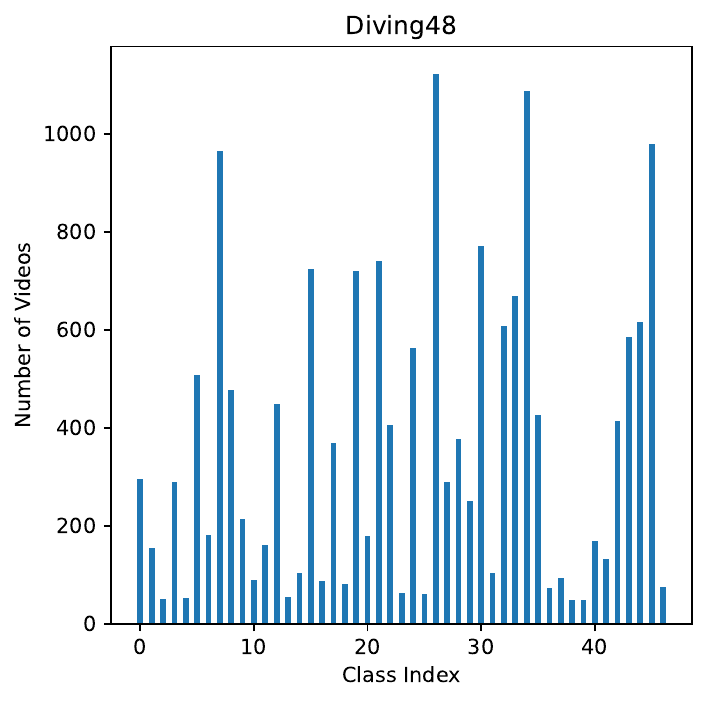}
	\end{subfigure}
	\begin{subfigure}{0.32\linewidth}
		\includegraphics[width=\linewidth]{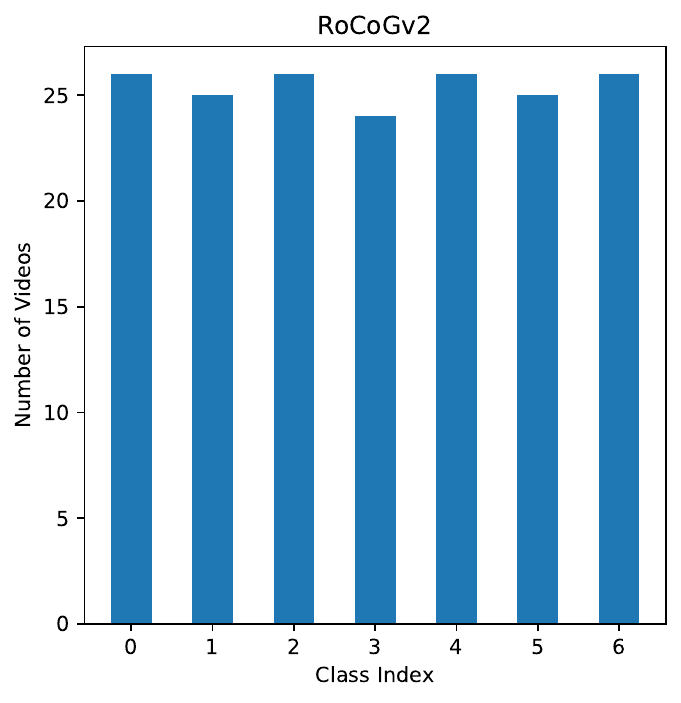}
	\end{subfigure}
	\caption{Class distributions for the 5 datasets used in this study.}
	\label{fig:class-distributions}
\end{figure}

\Cref{fig:class-distributions} shows the class distributions of each dataset used in this study. Here we witness a variety of distributions, where HMDB51, UCF101, and RoCoGv2 have a relatively balanced distribution (except one outlier in HMDB51), whereas SSv2 and Diving48 are unbalanced, with the number of videos per class ranging from 17 to 3170 for SSv2, and 50 to 1122 for Diving48. Based on \Cref{tab:dataset-summary-statistics}, we also notice the average number of videos per class varies greatly across all datasets, with an average of 25 videos per class for RoCoGv2, compared to 717 for SSv2. The effect that class distribution within the base dataset has on downstream few-shot performance has not been studied in this paper, nor any existing FSAR paper to date. We hypothesise that performance may not only depend on the difference between the base and novel dataset domains (as studied in this paper), but also the amount of training data available, class-wise and instance per class-wise. We leave understanding this aspect of base dataset design as future work.

\newpage
\section{Diving48 Splits}\label{app:diving_splits}

Here we list which classes within the Diving48 \cite{ferrari_resound_2018} we allocate to the training, validation, and testing sets for few-shot learning purposes. Classes were split randomly, with only portions specified. To do so, we combine the version 2 training and testing instance-wise splits proposed by the dataset creators (in which they remove one class from the original, hence only 47 classes), then randomly select 28, 9, and 10 classes to create the class-wise training, validation, and test splits, respectively.

\begin{minipage}[t]{.55\textwidth}
	\raggedright
	\underline{Training set classes (28):}
	\begin{itemize}
		\item Forward\_45som\_NoTwis\_TUCK
		\item Reverse\_35som\_NoTwis\_TUCK
		\item Reverse\_15som\_NoTwis\_PIKE
		\item Reverse\_25som\_NoTwis\_PIKE
		\item Reverse\_25som\_15Twis\_PIKE
		\item Back\_2som\_15Twis\_FREE
		\item Back\_2som\_25Twis\_FREE
		\item Inward\_15som\_NoTwis\_PIKE
		\item Reverse\_15som\_15Twis\_FREE
		\item Reverse\_Dive\_NoTwis\_TUCK
		\item Reverse\_15som\_35Twis\_FREE
		\item Reverse\_25som\_NoTwis\_TUCK
		\item Back\_3som\_NoTwis\_TUCK
		\item Reverse\_15som\_05Twis\_FREE
		\item Back\_25som\_NoTwis\_TUCK
		\item Back\_25som\_25Twis\_PIKE
		\item Back\_35som\_NoTwis\_TUCK
		\item Back\_15som\_NoTwis\_TUCK
		\item Forward\_Dive\_NoTwis\_PIKE
		\item Back\_15som\_NoTwis\_PIKE
		\item Forward\_25som\_2Twis\_PIKE
		\item Forward\_25som\_NoTwis\_TUCK
		\item Inward\_Dive\_NoTwis\_PIKE
		\item Back\_3som\_NoTwis\_PIKE
		\item Reverse\_15som\_25Twis\_FREE
		\item Back\_25som\_15Twis\_PIKE
		\item Back\_15som\_25Twis\_FREE
		\item Back\_15som\_05Twis\_FREE
	\end{itemize}
\end{minipage}
\begin{minipage}[t]{.45\textwidth}
	\raggedright
	
	\underline{Validation set classes (9):}
	\begin{itemize}
		\item Forward\_35som\_NoTwis\_TUCK
		\item Forward\_25som\_1Twis\_PIKE
		\item Forward\_25som\_NoTwis\_PIKE
		\item Inward\_25som\_NoTwis\_PIKE
		\item Back\_Dive\_NoTwis\_TUCK
		\item Back\_Dive\_NoTwis\_PIKE
		\item Inward\_15som\_NoTwis\_TUCK
		\item Forward\_1som\_NoTwis\_PIKE
		\item Reverse\_Dive\_NoTwis\_PIKE
	\end{itemize}
	\vspace{0.5cm}
	\underline{Testing set classes (10):}
	\begin{itemize}
		\item Forward\_25som\_3Twis\_PIKE
		\item Forward\_15som\_1Twis\_FREE
		\item Back\_15som\_15Twis\_FREE
		\item Forward\_35som\_NoTwis\_PIKE
		\item Inward\_25som\_NoTwis\_TUCK
		\item Back\_25som\_NoTwis\_PIKE
		\item Forward\_15som\_2Twis\_FREE
		\item Back\_35som\_NoTwis\_PIKE
		\item Inward\_35som\_NoTwis\_TUCK
		\item Forward\_15som\_NoTwis\_PIKE
	\end{itemize}
	
\end{minipage}

\newpage

\section{Domain Difference Measures}\label{app:domain-difference}

\begin{figure}[H]
	\includegraphics[width=0.85\textwidth]{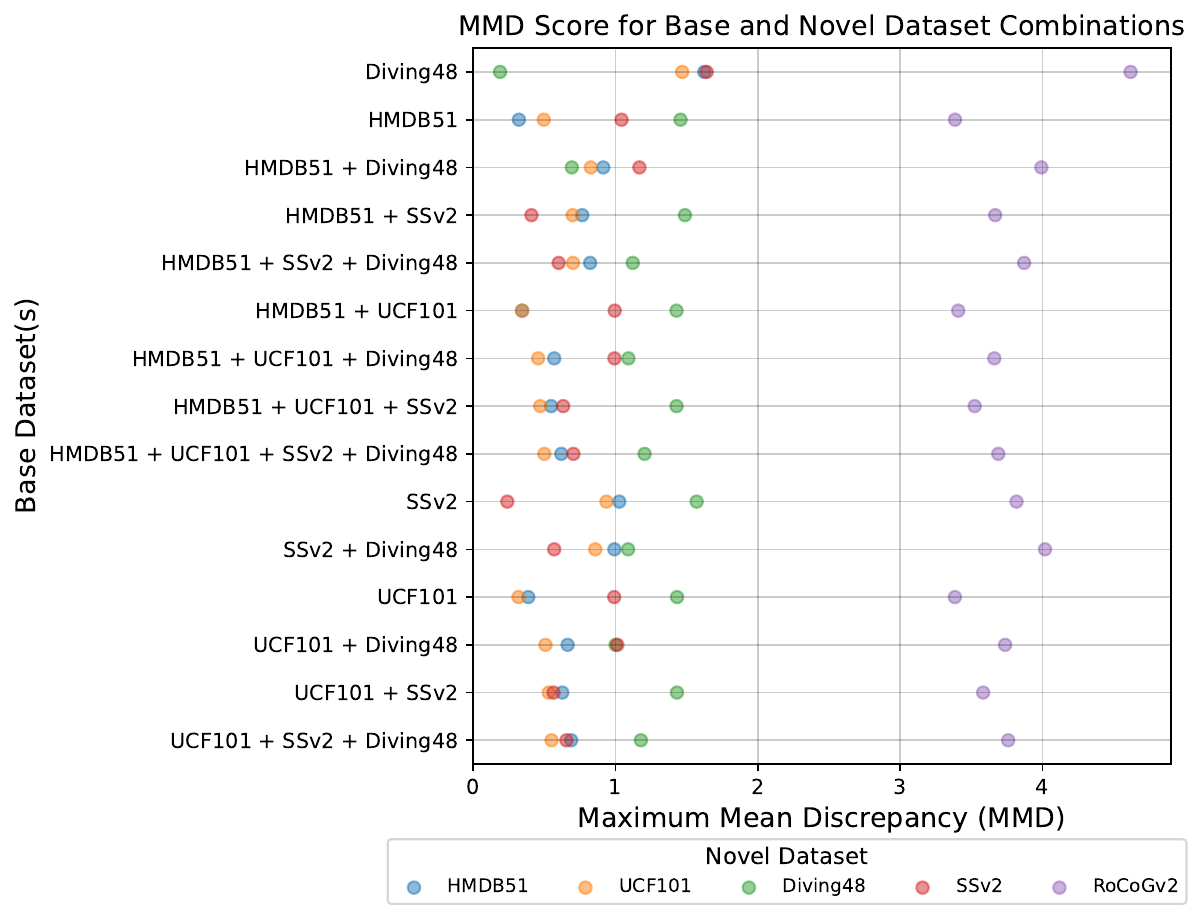}
	\caption{Domain difference scores (Maximum Mean Discrepancy) for all possible base and novel dataset combinations of the datasets chosen.}
	\label{fig:domain-difference-all}
\end{figure}

\Cref{fig:domain-difference-all} displays the Maximum Mean Discrepancy (MMD) scores calculated for all combinations of the datasets utilised for evaluation. Refer to \Cref{sec:measure} for details on how MMD is calculated. This visualisation assisted with choosing the combinations used for evaluation, in which combinations with the largest range and even distribution were chosen. Note that in comparison to the display style of \Cref{fig:domain-difference}, the y-axis and colour correspondence has switched (i.e., now the novel dataset is indicated by colour, and the base dataset is indicated on the y-axis). The results shown in \Cref{fig:domain-difference} are a subset of these results. 

\Cref{tab:raw-domain-difference-main} shows the raw MMD scores calculated for all combinations of the datasets used for evaluation. These are the raw values of those plotted in \Cref{fig:domain-difference-all}, and the bolded rows indicate the raw values for the combinations chosen for evaluation, which are plotted in \Cref{fig:domain-difference}.

\begin{table}[H]
	\tiny
	\centering
	\renewcommand{\arraystretch}{1.2}
	
	\caption{Summary of calculated average Maximum Mean Discrepancy (MMD) scores and 95\% confidence interval between each base (row) and novel (column) set. The bolded rows indicate which final combinations were chosen to use for evaluation.}
	\begin{tabularx}{\textwidth}{l *{5}{Y}}
		
		&
		\multicolumn{5}{c}{Novel Dataset}\\
		
		\Cline{2-6}{0.08em}
		
		
		& HMDB51
		& UCF101
		& Diving48
		& SSv2(-Full) 
		& RoCoGv2   \\
		
		
		\toprule
		
		\textbf{Diving48}
		& \textbf{1.623 ($\pm$ 0.17)}   
		& \textbf{1.468 ($\pm$ 0.17)}   
		& \textbf{0.189 ($\pm$ 0.08)}   
		& \textbf{1.640 ($\pm$ 0.15)}   
		& \textbf{4.618 ($\pm$ 0.14)}   
		\\
		
		
		\textbf{HMDB51}
		& \textbf{0.321 ($\pm$ 0.09)}   
		& \textbf{0.496 ($\pm$ 0.08)}   
		& \textbf{1.456 ($\pm$ 0.17)}   
		& \textbf{1.042 ($\pm$ 0.15)}   
		& \textbf{3.385 ($\pm$ 0.19)}   
		\\
		
		
		\textbf{HMDB51+Diving48}
		& \textbf{0.914 ($\pm$ 0.19)}   
		& \textbf{0.826 ($\pm$ 0.22)}   
		& \textbf{0.693 ($\pm$ 0.18)}   
		& \textbf{1.167 ($\pm$ 0.14)}   
		& \textbf{3.991 ($\pm$ 0.18)}   
		\\
		
		
		HMDB51+SSv2
		& 0.766 ($\pm$ 0.13)   
		& 0.699 ($\pm$ 0.13)   
		& 1.487 ($\pm$ 0.13)   
		& 0.409 ($\pm$ 0.12)   
		& 3.667 ($\pm$ 0.19)   
		\\
		
		
		HMDB51+SSv2+Diving48
		& 0.821 ($\pm$ 0.17)   
		& 0.701 ($\pm$ 0.20)   
		& 1.121 ($\pm$ 0.16)   
		& 0.600 ($\pm$ 0.12)   
		& 3.870 ($\pm$ 0.20)   
		\\
		
		
		HMDB51+UCF101
		& 0.344 ($\pm$ 0.09)   
		& 0.343 ($\pm$ 0.08)   
		& 1.428 ($\pm$ 0.20)   
		& 0.994 ($\pm$ 0.14)   
		& 3.407 ($\pm$ 0.16)   
		\\
		
		
		HMDB51+UCF101+Diving48
		& 0.569 ($\pm$ 0.15)   
		& 0.456 ($\pm$ 0.16)   
		& 1.090 ($\pm$ 0.20)   
		& 0.992 ($\pm$ 0.11)   
		& 3.661 ($\pm$ 0.21)   
		\\
		
		
		HMDB51+UCF101+SSv2
		& 0.547 ($\pm$ 0.14)   
		& 0.471 ($\pm$ 0.13)   
		& 1.428 ($\pm$ 0.17)   
		& 0.632 ($\pm$ 0.13)   
		& 3.523 ($\pm$ 0.18)   
		\\
		
		
		HMDB51+UCF101+SSv2+Diving48
		& 0.619 ($\pm$ 0.16)   
		& 0.500 ($\pm$ 0.18)   
		& 1.204 ($\pm$ 0.19)   
		& 0.704 ($\pm$ 0.13)   
		& 3.689 ($\pm$ 0.21)   
		\\
		
		
		SSv2
		& 1.026 ($\pm$ 0.11)   
		& 0.936 ($\pm$ 0.13)   
		& 1.569 ($\pm$ 0.11)   
		& 0.239 ($\pm$ 0.09)   
		& 3.817 ($\pm$ 0.19)   
		\\
		
		
		SSv2+Diving48
		& 0.992 ($\pm$ 0.17)   
		& 0.857 ($\pm$ 0.19)   
		& 1.088 ($\pm$ 0.16)   
		& 0.569 ($\pm$ 0.15)   
		& 4.017 ($\pm$ 0.19)   
		\\
		
		
		\textbf{UCF101}
		& \textbf{0.387 ($\pm$ 0.09)}   
		& \textbf{0.318 ($\pm$ 0.09)}   
		& \textbf{1.432 ($\pm$ 0.18)}   
		& \textbf{0.990 ($\pm$ 0.14)}   
		& \textbf{3.384 ($\pm$ 0.20)}   
		\\
		
		
		\textbf{UCF101+Diving48}
		& \textbf{0.663 ($\pm$ 0.17)}   
		& \textbf{0.507 ($\pm$ 0.17)}   
		& \textbf{1.000 ($\pm$ 0.21)}   
		& \textbf{1.012 ($\pm$ 0.10)}   
		& \textbf{3.736 ($\pm$ 0.18)}   
		\\
		
		
		UCF101+SSv2
		& 0.626 ($\pm$ 0.13)   
		& 0.530 ($\pm$ 0.13)   
		& 1.431 ($\pm$ 0.16)   
		& 0.564 ($\pm$ 0.14)   
		& 3.582 ($\pm$ 0.19)   
		\\
		
		
		UCF101+SSv2+Diving48
		& 0.688 ($\pm$ 0.18)   
		& 0.550 ($\pm$ 0.19)   
		& 1.178 ($\pm$ 0.18)   
		& 0.655 ($\pm$ 0.11)   
		& 3.758 ($\pm$ 0.19)   
		\\
		
		
		\bottomrule
		
	\end{tabularx}
	
	\label{tab:raw-domain-difference-main}
\end{table} 
\newpage
\section{Further Evaluation Setup Details}\label{app:further-eval-details}

\subsection{Further Model Descriptions}\label{app:models}

\textbf{ProtoNet} \cite{snell_prototypical_2017} is a simple, yet well established metric-based model which aims to learn a metric space in which classification can be performed by computing distances to prototype representations of each class. Typically, a prototype representation of a class is the mean value of the representations of the support set examples for each class. ResNet \cite{he_deep_2016} is used as the backbone, in which frame representations are averaged in feature space to obtain a video representation, and thus no explicit temporal modelling is used. We hyperparameter tune the learning rate for this model.

\textbf{STRM} \cite{thatipelli_spatio-temporal_2022} is currently the state-of-the-art single-domain few-shot action recognition model on HMDB51 and UCF101 datasets. Building upon TRX \cite{perrett_temporal-relational_2021} which uses a sub-sequence level self-attention mechanism for temporal modelling, STRM utilises a module which encodes patch-level (spatio) and frame-level (temporal) information into features. Specific to this model, we hyperparameter tune learning rate and number of episodes to train on.

\textbf{MoLo} \cite{wang_molo_2023} is currently the state-of-the-art single-domain few-shot action recognition model on the SSv2 dataset. This model uses a long-short contrastive objective to improve temporal context awareness during the matching process and a motion autodecoder to explicitly embed the feature network with motion dynamics. We hyperparameter tune the classification value and number of iterations between reducing the learning rate by a factor of 10.

\textbf{\textit{Transfer}} \cite{chen_closer_2019} is a commonly compared to baseline which consists of a feature extractor backbone and linear classification head. This model first undergoes fully supervised training using the base dataset, then the linear classification head is replaced during few-shot testing, in which the model is fine-tuned using the few labelled novel examples. The batch-size and epochs to reduce the learning rate by a factor of 10 are hyperparameter tuned.

\textbf{CDFSL-V }\cite{samarasinghe_cdfsl-v_2023} is a state-of-the-art cross-domain few-shot action recognition model whose pipeline consists of first leveraging unlabelled samples from both the base dataset and novel dataset to pre-train a VideoMAE \cite{tong_videomae_2022} autoencoder self-supervised, then performing curriculum learning to train the VideoMAE encoder further with the labelled base data and unlabelled novel data. For few-shot adaptation, a logistic regression classifier is learned on top of the encoder using the sampled 5-way 5-shot support set from the novel dataset.

\subsection{Further Implementation Details}

For STRM, MoLo, ProtoNet, and \textit{Transfer} models, a number of model parameters and training configurations are kept consistent for a fair comparison, whereby standard practices are employed. ResNet34 \cite{he_deep_2016} pre-trained on ImageNet \cite{deng_imagenet_2009} is used as the backbone feature extractor. To obtain the representation of a video, 8 frames are uniformly sampled, then augmentation includes rescaling frames to 256x256, then performing a 224x224 random or centre region crop during training and testing, respectively. Adam \cite{kingma_adam_2017} is used as the optimiser. Episodic models (STRM, MoLo, and ProtoNet) are trained under 5-way 5-shot tasks. Beyond these common configurations, refer to their respective papers for default training parameter values specific to each model, and refer to Tables \ref{tab:ProtoNet-hyperparams} to \ref{tab:Transfer-hyperparams} for the hyperparameter values which were found after tuning. For these models, 5 models each were trained, each using a different base dataset combination.

For CDFSL-V, we follow exactly the implementation and training details outlined in their paper \cite{samarasinghe_cdfsl-v_2023}. Since this model leverages unlabelled novel data during training, one model is trained for each of the 25 scenarios.

\subsection{Tuned Hyperparameters}

Tables \ref{tab:ProtoNet-hyperparams} to \ref{tab:Transfer-hyperparams} below indicate the Hyperparameters that were tuned for each model, and the resultant value used to train the final model used for evaluation. All other model hyperparameters were kept consistent with those found to be most effective in the ablation studies conducted in the original papers, or kept as the default values provided in the source code for each model. As seen in the tables, 5 versions of each model were trained, each utilising different base data during training.

\begin{table}[H]
	\centering
	\footnotesize
		\renewcommand{\arraystretch}{1.1}
	
	\caption{The resultant hyperparameter value used to train each ProtoNet model, where each model is trained using different base data.}
	
	\begin{tabularx}{\textwidth}{l *{1}{Y}}
		
		&
		\multicolumn{1}{c}{\textit{Hyperparameter}}\\
		
		\Cline{2-2}{0.08em}
		
		\textit{Base Dataset}
		& Learning Rate
		\\
		
		\toprule
		
		
		HMDB51
		& 1e-05
		\\
		
		UCF101
		& 1e-05
		\\
		
		Diving48
		& 1e-05
		\\
		
		Diving48 + HMDB51
		& 1e-04
		\\
		
		Diving48 + UCF101
		& 1e-04
		\\
		
		\bottomrule
	\end{tabularx}

	\label{tab:ProtoNet-hyperparams}
\end{table}

\begin{table}[H]
	\centering
	\footnotesize
		\renewcommand{\arraystretch}{1.1}
	
	\caption{The resultant hyperparameter values used to train each STRM model, where each model is trained using different base data.}
	
\begin{tabularx}{\textwidth}{l *{2}{Y}}
	
	&
	\multicolumn{2}{c}{\textit{Hyperparameter}}\\
	
	\Cline{2-3}{0.08em}
	
	\textit{Base Dataset}
	& Learning Rate
	& No. Training Episodes
	\\
	
	\toprule
	
	
	HMDB51
	& 1e-05
	& 44000
	\\
	
	UCF101
	& 1e-05
	& 99000
	\\
	
	Diving48
	& 1e-04
	& 84000
	\\
	
	Diving48 + HMDB51
	& 1e-04
	& 47000
	\\
	
	Diving48 + UCF101
	& 1e-04
	& 94000
	\\
	
	\bottomrule
\end{tabularx}

	\label{tab:STRM-hyperparams}
\end{table} 

\begin{table}[H]
	\centering
	\footnotesize
		\renewcommand{\arraystretch}{1.1}
	
	\caption{The resultant hyperparameter values used to train each MoLo model, where each model is trained using different base data.}
	
\begin{tabularx}{\textwidth}{l *{3}{Y}}
	
	&
	\multicolumn{3}{c}{\textit{Hyperparameter}}\\
	
	\Cline{2-4}{0.08em}
	
	\textit{Base Dataset}
	& Classification Value
	& Iterations per step
	& No. Training Episodes
	\\
	
	\toprule
	
	
	HMDB51
	& 0.6
	& 300
	& 1680
	\\
	
	UCF101
	& 0.3
	& 300
	& 1260
	\\
	
	Diving48
	& 0.6
	& 1200
	& 10560
	\\
	
	Diving48 + HMDB51
	& 0.6
	& 1200
	& 6720
	\\
	
	Diving48 + UCF101
	& 0.6
	& 1200
	& 6000
	\\
	
	\bottomrule
\end{tabularx}

	\label{tab:MoLo-hyperparams}
\end{table} 

\begin{table}[H]
	\centering
	\footnotesize
		\renewcommand{\arraystretch}{1.1}
	
	\caption{The resultant hyperparameter values used to train the backbone of each \textit{Transfer} model, where each model is trained using different base data. `LR Adjustment Steps' refers to the epochs in which the learning rate is decreased by a factor of 10.}
	\begin{tabularx}{\textwidth}{l *{2}{Y}}
		
		&
		\multicolumn{2}{c}{\textit{Hyperparameter}}\\
		
		\Cline{2-3}{0.08em}
		
		\textit{Base Dataset}
		& Batchsize
		& LR Adjustment steps
		\\
		
		\toprule
		
		
		HMDB51
		& 32
		& 25, 50
		\\
		
		UCF101
		& 32
		& 25, 50
		\\
		
		Diving48
		& 32
		& 25, 50
		\\
		
		Diving48 + HMDB51
		& 32
		& 25, 50
		\\
		
		Diving48 + UCF101
		& 32
		& 25, 50
		\\
		
		\bottomrule
	\end{tabularx}

	\label{tab:Transfer-hyperparams}
\end{table} 
\newpage
\section{5-Way 5-Shot Results}\label{app:raw_5shot_results}

\begin{table}[H]
	\centering
	\scriptsize
	
	\caption{5-Way 5-Shot performances of each model when evaluated on different novel datasets. Each column indicates which base dataset the model was trained on, however last column indicates an untrained model, with random weights. The average performance and 95\% confidence interval is shown, after evaluating on 10000 randomly sampled episodes. These are the numeric results for those shown in \Cref{fig:model-fs-performance}}
	\vspace{0.5cm}

\begin{tabularx}{\textwidth}{l *{6}{Y}}
	
	\multicolumn{7}{c}{\textit{Novel Dataset}: HMDB51}\\
	
	\toprule
	
	\multirow{2}{*}{\textit{Base Dataset:}}
	& \multirow{2}{*}{HMDB51}
	& \multirow{2}{*}{UCF101}
	& \multirow{2}{*}{Diving48}
	& Diving48 
	& Diving48 
	& Random   \\
	& 
	& 
	& 
	& + HMDB51
	& + UCF101
	& Weights  \\
	
	\toprule
	
	
	ProtoNet \cite{snell_prototypical_2017} 
	&  65.9 ($\pm$ 0.41)  
	&  64.4 ($\pm$ 0.40)  
	&  57.6 ($\pm$ 0.43)  
	&  57.7 ($\pm$ 0.42)  
	&  58.3 ($\pm$ 0.41)  
	&  27.3 ($\pm$ 0.35)  
	\\
	
	STRM \cite{thatipelli_spatio-temporal_2022} 
	&  67.8 ($\pm$ 0.40)  
	&  67.2 ($\pm$ 0.40)  
	&  40.3 ($\pm$ 0.42)  
	&  61.0 ($\pm$ 0.41)  
	&  58.9 ($\pm$ 0.42)  
	&  27.8 ($\pm$ 0.36)  
	\\
	
	MoLo \cite{wang_molo_2023} 
	&  69.9 ($\pm$ 0.39)  
	&  64.9 ($\pm$ 0.40)  
	&  43.8 ($\pm$ 0.41)  
	&  65.2 ($\pm$ 0.40)  
	&  59.7 ($\pm$ 0.41)  
	&  26.6 ($\pm$ 0.38)  
	\\
	
	\textit{Transfer} \cite{zhu_closer_2021} 
	&  53.7 ($\pm$ 0.43)  
	&  48.8 ($\pm$ 0.42)  
	&  42.6 ($\pm$ 0.43)  
	&  49.1 ($\pm$ 0.43)  
	&  45.8 ($\pm$ 0.42)  
	&  21.7 ($\pm$ 0.21)  
	\\
	
	CDFSL-V \cite{samarasinghe_cdfsl-v_2023} 
	&  30.8 ($\pm$ 0.61)  
	&  36.0 ($\pm$ 0.20)  
	&  27.0 ($\pm$ 0.58)  
	&  32.0 ($\pm$ 0.19)  
	&  34.5 ($\pm$ 0.19)  
	&  30.7 ($\pm$ 0.19)  
	\\
	
	\bottomrule
	& 
	& 
	& 
	& 
	& 
	& 
	\\
	
\end{tabularx}

\begin{tabularx}{\textwidth}{l *{6}{Y}}
	
	\multicolumn{7}{c}{\textit{Novel Dataset}: UCF101}\\
	
	\toprule
	
	\multirow{2}{*}{\textit{Base Dataset:}}
	& \multirow{2}{*}{HMDB51}
	& \multirow{2}{*}{UCF101}
	& \multirow{2}{*}{Diving48}
	& Diving48 
	& Diving48 
	& Random   \\
	& 
	& 
	& 
	& + HMDB51
	& + UCF101
	& Weights  \\
	
	\toprule
	
	
	ProtoNet \cite{snell_prototypical_2017} 
	&  90.9 ($\pm$ 0.25)  
	&  92.3 ($\pm$ 0.23)  
	&  84.6 ($\pm$ 0.31)  
	&  84.2 ($\pm$ 0.32)  
	&  85.6 ($\pm$ 0.30)  
	&  31.4 ($\pm$ 0.38)  
	\\
	
	STRM \cite{thatipelli_spatio-temporal_2022} 
	&  91.2 ($\pm$ 0.25)  
	&  94.3 ($\pm$ 0.20)  
	&  70.0 ($\pm$ 0.40)  
	&  84.3 ($\pm$ 0.32)  
	&  87.4 ($\pm$ 0.29)  
	&  32.9 ($\pm$ 0.39)  
	\\
	
	MoLo \cite{wang_molo_2023} 
	&  89.6 ($\pm$ 0.27)  
	&  92.9 ($\pm$ 0.22)  
	&  71.9 ($\pm$ 0.39)  
	&  87.3 ($\pm$ 0.30)  
	&  92.0 ($\pm$ 0.24)  
	&  41.5 ($\pm$ 0.42)  
	\\
	
	\textit{Transfer} \cite{zhu_closer_2021} 
	&  82.2 ($\pm$ 0.33)  
	&  80.6 ($\pm$ 0.34)  
	&  71.2 ($\pm$ 0.39)  
	&  78.6 ($\pm$ 0.36)  
	&  76.4 ($\pm$ 0.37)  
	&  26.4 ($\pm$ 0.25)  
	\\
	
	CDFSL-V \cite{samarasinghe_cdfsl-v_2023} 
	&  46.3 ($\pm$ 0.23)  
	&  56.1 ($\pm$ 0.23)  
	&  41.1 ($\pm$ 0.22)  
	&  47.7 ($\pm$ 0.23)  
	&  56.2 ($\pm$ 0.23)  
	&  29.1 ($\pm$ 0.23)  
	\\
	
	\bottomrule
	& 
	& 
	& 
	& 
	& 
	& 
	\\
	
\end{tabularx}

\begin{tabularx}{\textwidth}{l *{6}{Y}}
	
	\multicolumn{7}{c}{\textit{Novel Dataset}: Diving48}\\
	
	\toprule
	
	\multirow{2}{*}{\textit{Base Dataset:}}
	& \multirow{2}{*}{HMDB51}
	& \multirow{2}{*}{UCF101}
	& \multirow{2}{*}{Diving48}
	& Diving48 
	& Diving48 
	& Random   \\
	& 
	& 
	& 
	& + HMDB51
	& + UCF101
	& Weights  \\
	
	\toprule
	
	
	ProtoNet \cite{snell_prototypical_2017} 
	&  34.7 ($\pm$ 0.41)  
	&  35.6 ($\pm$ 0.41)  
	&  64.7 ($\pm$ 0.42)  
	&  56.4 ($\pm$ 0.43)  
	&  57.6 ($\pm$ 0.42)  
	&  21.2 ($\pm$ 0.33)  
	\\
	
	STRM \cite{thatipelli_spatio-temporal_2022} 
	&  37.5 ($\pm$ 0.42)  
	&  37.3 ($\pm$ 0.41)  
	&  79.5 ($\pm$ 0.35)  
	&  71.6 ($\pm$ 0.39)  
	&  75.6 ($\pm$ 0.37)  
	&  22.1 ($\pm$ 0.34)  
	\\
	
	MoLo \cite{wang_molo_2023} 
	&  36.7 ($\pm$ 0.41)  
	&  34.9 ($\pm$ 0.41)  
	&  83.7 ($\pm$ 0.32)  
	&  79.0 ($\pm$ 0.35)  
	&  76.9 ($\pm$ 0.36)  
	&  24.4 ($\pm$ 0.37)  
	\\
	
	\textit{Transfer} \cite{zhu_closer_2021} 
	&  36.2 ($\pm$ 0.41)  
	&  35.6 ($\pm$ 0.41)  
	&  61.5 ($\pm$ 0.42)  
	&  60.9 ($\pm$ 0.42)  
	&  57.7 ($\pm$ 0.42)  
	&  20.5 ($\pm$ 0.14)  
	\\
	
	CDFSL-V \cite{samarasinghe_cdfsl-v_2023} 
	&  26.2 ($\pm$ 0.18)  
	&  26.8 ($\pm$ 0.17)  
	&  22.9 ($\pm$ 0.17)  
	&  34.5 ($\pm$ 0.20)  
	&  35.8 ($\pm$ 0.20)  
	&  27.3 ($\pm$ 0.18)  
	\\
	
	\bottomrule
	& 
	& 
	& 
	& 
	& 
	& 
	\\
	
\end{tabularx}

\begin{tabularx}{\textwidth}{l *{6}{Y}}
	
	\multicolumn{7}{c}{\textit{Novel Dataset}: SSv2}\\
	
	\toprule
	
	\multirow{2}{*}{\textit{Base Dataset:}}
	& \multirow{2}{*}{HMDB51}
	& \multirow{2}{*}{UCF101}
	& \multirow{2}{*}{Diving48}
	& Diving48 
	& Diving48 
	& Random   \\
	& 
	& 
	& 
	& + HMDB51
	& + UCF101
	& Weights  \\
	
	\toprule
	
	
	ProtoNet \cite{snell_prototypical_2017} 
	&  37.7 ($\pm$ 0.41)  
	&  38.1 ($\pm$ 0.42)  
	&  34.8 ($\pm$ 0.41)  
	&  34.3 ($\pm$ 0.41)  
	&  36.3 ($\pm$ 0.42)  
	&  21.9 ($\pm$ 0.34)  
	\\
	
	STRM \cite{thatipelli_spatio-temporal_2022} 
	&  38.6 ($\pm$ 0.42)  
	&  40.5 ($\pm$ 0.43)  
	&  28.6 ($\pm$ 0.39)  
	&  33.2 ($\pm$ 0.41)  
	&  34.7 ($\pm$ 0.41)  
	&  22.4 ($\pm$ 0.34)  
	\\
	
	MoLo \cite{wang_molo_2023} 
	&  37.4 ($\pm$ 0.42)  
	&  39.2 ($\pm$ 0.42)  
	&  29.1 ($\pm$ 0.39)  
	&  33.0 ($\pm$ 0.40)  
	&  36.4 ($\pm$ 0.41)  
	&  23.0 ($\pm$ 0.36)  
	\\
	
	\textit{Transfer} \cite{zhu_closer_2021} 
	&  34.8 ($\pm$ 0.41)  
	&  33.7 ($\pm$ 0.40)  
	&  30.7 ($\pm$ 0.40)  
	&  33.4 ($\pm$ 0.41)  
	&  32.7 ($\pm$ 0.40)  
	&  20.6 ($\pm$ 0.20)  
	\\
	
	CDFSL-V \cite{samarasinghe_cdfsl-v_2023} 
	&  21.5 ($\pm$ 0.51)  
	&  24.7 ($\pm$ 0.56)  
	&  20.9 ($\pm$ 0.49)  
	&  22.5 ($\pm$ 0.53)  
	&  23.6 ($\pm$ 0.17)  
	&  22.9 ($\pm$ 0.17)  
	\\
	
	\bottomrule
	& 
	& 
	& 
	& 
	& 
	& 
	\\
	
\end{tabularx}

\begin{tabularx}{\textwidth}{l *{6}{Y}}
	
	\multicolumn{7}{c}{\textit{Novel Dataset}: RoCoGv2}\\
	
	\toprule
	
	\multirow{2}{*}{\textit{Base Dataset:}}
	& \multirow{2}{*}{HMDB51}
	& \multirow{2}{*}{UCF101}
	& \multirow{2}{*}{Diving48}
	& Diving48 
	& Diving48 
	& Random   \\
	& 
	& 
	& 
	& + HMDB51
	& + UCF101
	& Weights  \\
	
	\toprule
	
	
	ProtoNet \cite{snell_prototypical_2017} 
	&  23.8 ($\pm$ 0.37)  
	&  22.1 ($\pm$ 0.36)  
	&  23.0 ($\pm$ 0.37)  
	&  32.5 ($\pm$ 0.41)  
	&  34.5 ($\pm$ 0.42)  
	&  19.0 ($\pm$ 0.32)  
	\\
	
	STRM \cite{thatipelli_spatio-temporal_2022} 
	&  30.9 ($\pm$ 0.41)  
	&  36.0 ($\pm$ 0.43)  
	&  30.4 ($\pm$ 0.40)  
	&  30.6 ($\pm$ 0.40)  
	&  31.7 ($\pm$ 0.40)  
	&  19.2 ($\pm$ 0.32)  
	\\
	
	MoLo \cite{wang_molo_2023} 
	&  24.6 ($\pm$ 0.37)  
	&  22.9 ($\pm$ 0.36)  
	&  27.3 ($\pm$ 0.39)  
	&  25.8 ($\pm$ 0.38)  
	&  26.4 ($\pm$ 0.38)  
	&  19.3 ($\pm$ 0.34)  
	\\
	
	\textit{Transfer} \cite{zhu_closer_2021} 
	&  48.1 ($\pm$ 0.43)  
	&  47.4 ($\pm$ 0.44)  
	&  46.3 ($\pm$ 0.44)  
	&  47.8 ($\pm$ 0.43)  
	&  46.4 ($\pm$ 0.43)  
	&  20.0 ($\pm$ 0.07)  
	\\
	
	CDFSL-V \cite{samarasinghe_cdfsl-v_2023} 
	&  23.9 ($\pm$ 0.50)  
	&  25.3 ($\pm$ 0.52)  
	&  21.5 ($\pm$ 0.52)  
	&  27.3 ($\pm$ 0.56)  
	&  27.6 ($\pm$ 0.55)  
	&  19.2 ($\pm$ 0.15)  
	\\
	
	\bottomrule
	& 
	& 
	& 
	& 
	& 
	& 
	\\
	
\end{tabularx}

		\label{tab:5shot-results}
	\end{table} 

\newpage
\section{5-Way 1-Shot Results}\label{app:1-shot-results}
\begin{figure}[H]
	\centering
	\includegraphics[width=0.9\linewidth]{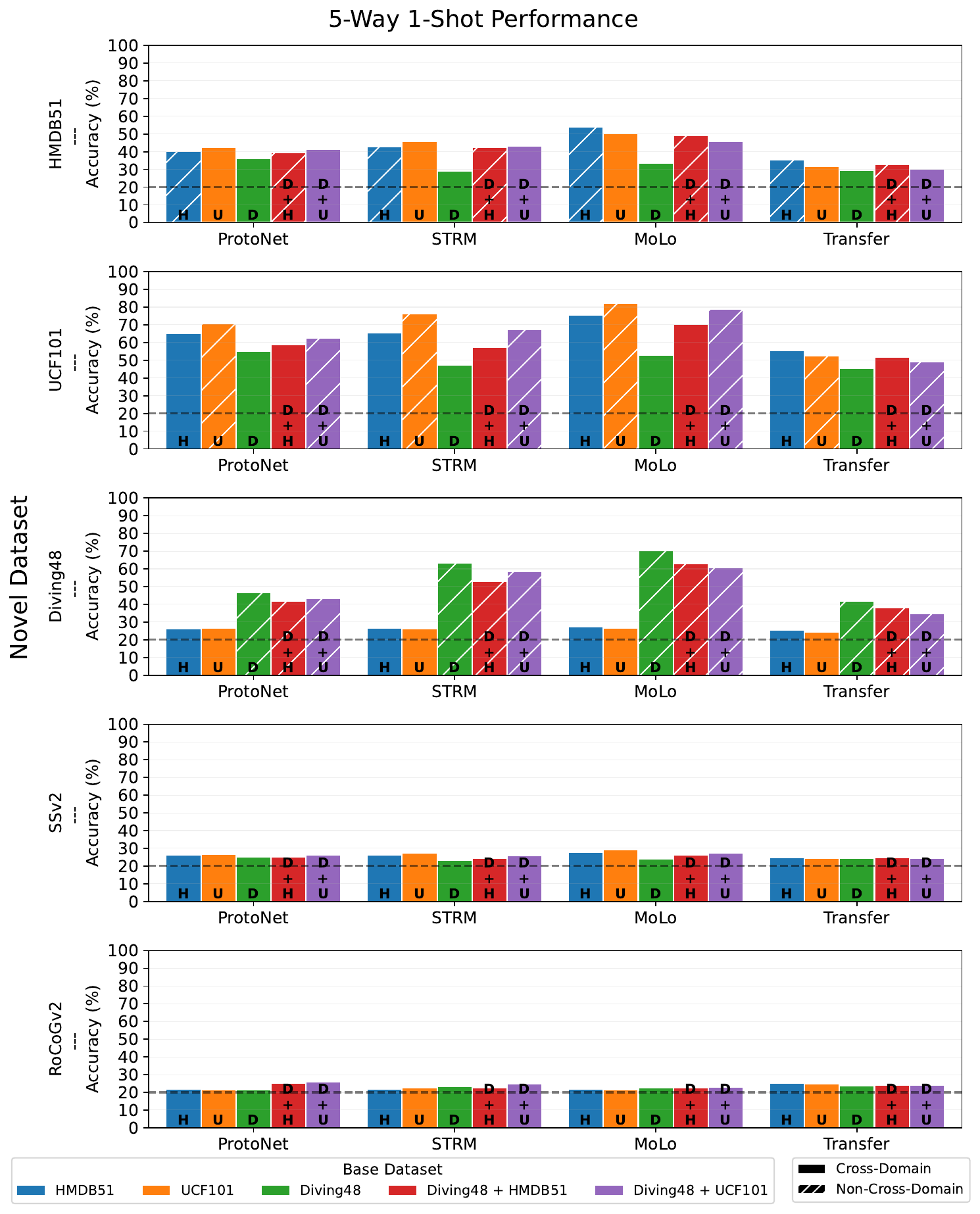}
	\caption{5-Way 1-Shot performance of existing single-domain (ProtoNet, STRM, MoLo) and transfer-based (\textit{Transfer}) few-shot action recognition models trained using different base datasets (indicated by colour), and evaluated on different novel datasets (each subfigure). Hatching indicates which combinations are not true cross-domain scenarios. The dotted line at 20\% indicates the random choice likelihood. The raw values are found in \Cref{tab:1shot_results}.}
	\label{fig:model-fs-performance-1shot}
\end{figure}

We evaluated each model under 5-way 1-shot settings. We did not re-evaluate CDFSL-V under this setting as it requires learning a logistic regression classifier, which is unsuitable for 1-shot tasks. The visualisation of these results is shown in \Cref{fig:model-fs-performance-1shot}. The raw values are shown in \Cref{tab:1shot_results}. 

These results show similar trends to those analysed in the main body of the paper for 5-way 5-shot performance. First, ProtoNet, an earlier single-domain method which does not use temporal alignment, performs comparably and in some instances better than the newer, SOTA single-domain methods, STRM and MoLo, which leverage temporal alignment. This again suggests that existing SOTA temporal alignment techniques fail to generalise on unseen domains. Under cross-domain settings (i.e., solid bars), we again witness different models performing the same, better, and worse when data is added to the base set, supporting the original findings of a universal sensitivity to base class construction which needs further exploration. We also observe that when the base and novel sets contain samples from the same domain (i.e., hatched bars), performance is boosted, especially for single-domain models (ProtoNet, STRM, and MoLo), demonstrating the bias introduced when learning from data sampled from the same domain used for model evaluation. 

Contradictory to the main results, the performance on RoCoGv2 and SSv2 novel sets is consistently poor across all models, including the transfer-based approach. This suggests that one example per novel class is not sufficient enough to transfer domain knowledge during fine-tuning, and thus a minimum number of shots may need to be placed for benchmarking cross-domain methods. 

\begin{table}[H]
\centering
\footnotesize

\caption{5-Way 1-Shot performances of each model when evaluated on different novel datasets. Each column indicates which base dataset the model was trained on. The average performance and 95\% confidence interval is shown, after evaluating on 10000 randomly sampled 5-Way 1-Shot episodes. These are the numeric values of those shown in \Cref{fig:domain-diff-vs-fewshot-performance-1shot}}
\vspace{0.5cm}
\begin{tabularx}{\textwidth}{l *{5}{Y}}
	
	\multicolumn{6}{c}{\textit{Novel Dataset}: HMDB51}\\
	
	\toprule
	
	\multirow{2}{*}{\textit{Base Dataset:}}
	& \multirow{2}{*}{HMDB51}
	& \multirow{2}{*}{UCF101}
	& \multirow{2}{*}{Diving48}
	& Diving48 
	& Diving48   \\
	& 
	& 
	& 
	& + HMDB51
	& + UCF101  \\
	
	\toprule
	
	
	ProtoNet \cite{snell_prototypical_2017} 
	&  40.4 ($\pm$ 0.38)  
	&  42.4 ($\pm$ 0.39)  
	&  36.1 ($\pm$ 0.38)  
	&  39.4 ($\pm$ 0.39)  
	&  41.4 ($\pm$ 0.39)  
	\\
	
	STRM \cite{thatipelli_spatio-temporal_2022} 
	&  42.9 ($\pm$ 0.39)  
	&  45.8 ($\pm$ 0.40)  
	&  29.2 ($\pm$ 0.37)  
	&  42.6 ($\pm$ 0.40)  
	&  43.4 ($\pm$ 0.40)  
	\\
	
	MoLo \cite{wang_molo_2023} 
	&  54.2 ($\pm$ 0.42)  
	&  50.1 ($\pm$ 0.41)  
	&  33.5 ($\pm$ 0.40)  
	&  49.2 ($\pm$ 0.42)  
	&  45.9 ($\pm$ 0.40)  
	\\
	
	\textit{Transfer} \cite{zhu_closer_2021} 
	&  35.4 ($\pm$ 0.36)  
	&  31.7 ($\pm$ 0.35)  
	&  29.7 ($\pm$ 0.37)  
	&  32.8 ($\pm$ 0.37)  
	&  30.4 ($\pm$ 0.35)  
	\\
	
	\bottomrule
	& 
	& 
	& 
	& 
	& 
	\\
	
\end{tabularx}

\begin{tabularx}{\textwidth}{l *{5}{Y}}
	
	\multicolumn{6}{c}{\textit{Novel Dataset}: UCF101}\\
	
	\toprule
	
	\multirow{2}{*}{\textit{Base Dataset:}}
	& \multirow{2}{*}{HMDB51}
	& \multirow{2}{*}{UCF101}
	& \multirow{2}{*}{Diving48}
	& Diving48 
	& Diving48   \\
	& 
	& 
	& 
	& + HMDB51
	& + UCF101  \\
	
	\toprule
	
	
	ProtoNet \cite{snell_prototypical_2017} 
	&  65.1 ($\pm$ 0.39)  
	&  70.8 ($\pm$ 0.38)  
	&  55.2 ($\pm$ 0.40)  
	&  58.9 ($\pm$ 0.40)  
	&  62.4 ($\pm$ 0.39)  
	\\
	
	STRM \cite{thatipelli_spatio-temporal_2022} 
	&  65.5 ($\pm$ 0.39)  
	&  76.3 ($\pm$ 0.35)  
	&  47.3 ($\pm$ 0.41)  
	&  57.6 ($\pm$ 0.40)  
	&  67.4 ($\pm$ 0.39)  
	\\
	
	MoLo \cite{wang_molo_2023} 
	&  75.6 ($\pm$ 0.37)  
	&  82.1 ($\pm$ 0.33)  
	&  52.8 ($\pm$ 0.43)  
	&  70.5 ($\pm$ 0.40)  
	&  78.8 ($\pm$ 0.35)  
	\\
	
	\textit{Transfer} \cite{zhu_closer_2021} 
	&  55.4 ($\pm$ 0.39)  
	&  52.7 ($\pm$ 0.39)  
	&  45.4 ($\pm$ 0.40)  
	&  51.7 ($\pm$ 0.39)  
	&  49.3 ($\pm$ 0.39)  
	\\
	
	\bottomrule
	& 
	& 
	& 
	& 
	& 
	\\
	
\end{tabularx}

\begin{tabularx}{\textwidth}{l *{5}{Y}}
	
	\multicolumn{6}{c}{\textit{Novel Dataset}: Diving48}\\
	
	\toprule
	
	\multirow{2}{*}{\textit{Base Dataset:}}
	& \multirow{2}{*}{HMDB51}
	& \multirow{2}{*}{UCF101}
	& \multirow{2}{*}{Diving48}
	& Diving48 
	& Diving48   \\
	& 
	& 
	& 
	& + HMDB51
	& + UCF101  \\
	
	\toprule
	
	
	ProtoNet \cite{snell_prototypical_2017} 
	&  26.1 ($\pm$ 0.35)  
	&  26.5 ($\pm$ 0.36)  
	&  46.5 ($\pm$ 0.40)  
	&  41.9 ($\pm$ 0.41)  
	&  43.2 ($\pm$ 0.40)  
	\\
	
	STRM \cite{thatipelli_spatio-temporal_2022} 
	&  26.7 ($\pm$ 0.35)  
	&  26.4 ($\pm$ 0.35)  
	&  63.3 ($\pm$ 0.40)  
	&  52.9 ($\pm$ 0.40)  
	&  58.6 ($\pm$ 0.41)  
	\\
	
	MoLo \cite{wang_molo_2023} 
	&  27.4 ($\pm$ 0.37)  
	&  26.6 ($\pm$ 0.37)  
	&  70.3 ($\pm$ 0.38)  
	&  62.8 ($\pm$ 0.40)  
	&  60.6 ($\pm$ 0.40)  
	\\
	
	\textit{Transfer} \cite{zhu_closer_2021} 
	&  25.4 ($\pm$ 0.34)  
	&  24.6 ($\pm$ 0.33)  
	&  41.8 ($\pm$ 0.40)  
	&  38.3 ($\pm$ 0.39)  
	&  34.7 ($\pm$ 0.37)  
	\\
	
	\bottomrule
	& 
	& 
	& 
	& 
	& 
	\\
	
\end{tabularx}

\begin{tabularx}{\textwidth}{l *{5}{Y}}
	
	\multicolumn{6}{c}{\textit{Novel Dataset}: SSv2}\\
	
	\toprule
	
	\multirow{2}{*}{\textit{Base Dataset:}}
	& \multirow{2}{*}{HMDB51}
	& \multirow{2}{*}{UCF101}
	& \multirow{2}{*}{Diving48}
	& Diving48 
	& Diving48   \\
	& 
	& 
	& 
	& + HMDB51
	& + UCF101  \\
	
	\toprule
	
	
	ProtoNet \cite{snell_prototypical_2017} 
	&  26.2 ($\pm$ 0.32)  
	&  26.5 ($\pm$ 0.33)  
	&  25.2 ($\pm$ 0.32)  
	&  25.2 ($\pm$ 0.34)  
	&  26.3 ($\pm$ 0.34)  
	\\
	
	STRM \cite{thatipelli_spatio-temporal_2022} 
	&  26.3 ($\pm$ 0.33)  
	&  27.4 ($\pm$ 0.34)  
	&  23.3 ($\pm$ 0.34)  
	&  24.4 ($\pm$ 0.33)  
	&  25.9 ($\pm$ 0.35)  
	\\
	
	MoLo \cite{wang_molo_2023} 
	&  27.9 ($\pm$ 0.38)  
	&  29.2 ($\pm$ 0.38)  
	&  24.2 ($\pm$ 0.36)  
	&  26.2 ($\pm$ 0.37)  
	&  27.3 ($\pm$ 0.38)  
	\\
	
	\textit{Transfer} \cite{zhu_closer_2021} 
	&  24.9 ($\pm$ 0.31)  
	&  24.6 ($\pm$ 0.30)  
	&  24.3 ($\pm$ 0.34)  
	&  24.8 ($\pm$ 0.32)  
	&  24.5 ($\pm$ 0.32)  
	\\
	
	\bottomrule
	& 
	& 
	& 
	& 
	& 
	\\
	
\end{tabularx}

\begin{tabularx}{\textwidth}{l *{5}{Y}}
	
	\multicolumn{6}{c}{\textit{Novel Dataset}: RoCoGv2}\\
	
	\toprule
	
	\multirow{2}{*}{\textit{Base Dataset:}}
	& \multirow{2}{*}{HMDB51}
	& \multirow{2}{*}{UCF101}
	& \multirow{2}{*}{Diving48}
	& Diving48 
	& Diving48   \\
	& 
	& 
	& 
	& + HMDB51
	& + UCF101  \\
	
	\toprule
	
	
	ProtoNet \cite{snell_prototypical_2017} 
	&  22.0 ($\pm$ 0.35)  
	&  21.3 ($\pm$ 0.34)  
	&  21.3 ($\pm$ 0.34)  
	&  25.2 ($\pm$ 0.36)  
	&  25.9 ($\pm$ 0.37)  
	\\
	
	STRM \cite{thatipelli_spatio-temporal_2022} 
	&  21.9 ($\pm$ 0.35)  
	&  22.8 ($\pm$ 0.36)  
	&  23.2 ($\pm$ 0.35)  
	&  22.5 ($\pm$ 0.36)  
	&  24.7 ($\pm$ 0.37)  
	\\
	
	MoLo \cite{wang_molo_2023} 
	&  21.8 ($\pm$ 0.35)  
	&  21.5 ($\pm$ 0.34)  
	&  22.6 ($\pm$ 0.36)  
	&  22.5 ($\pm$ 0.36)  
	&  22.9 ($\pm$ 0.35)  
	\\
	
	\textit{Transfer} \cite{zhu_closer_2021} 
	&  25.4 ($\pm$ 0.36)  
	&  24.8 ($\pm$ 0.35)  
	&  23.6 ($\pm$ 0.36)  
	&  24.0 ($\pm$ 0.36)  
	&  24.1 ($\pm$ 0.34)  
	\\
	
	\bottomrule
	& 
	& 
	& 
	& 
	& 
	\\
	
\end{tabularx}

\label{tab:1shot_results}
\end{table} 

\begin{figure}[h]
	\includegraphics[width=\textwidth]{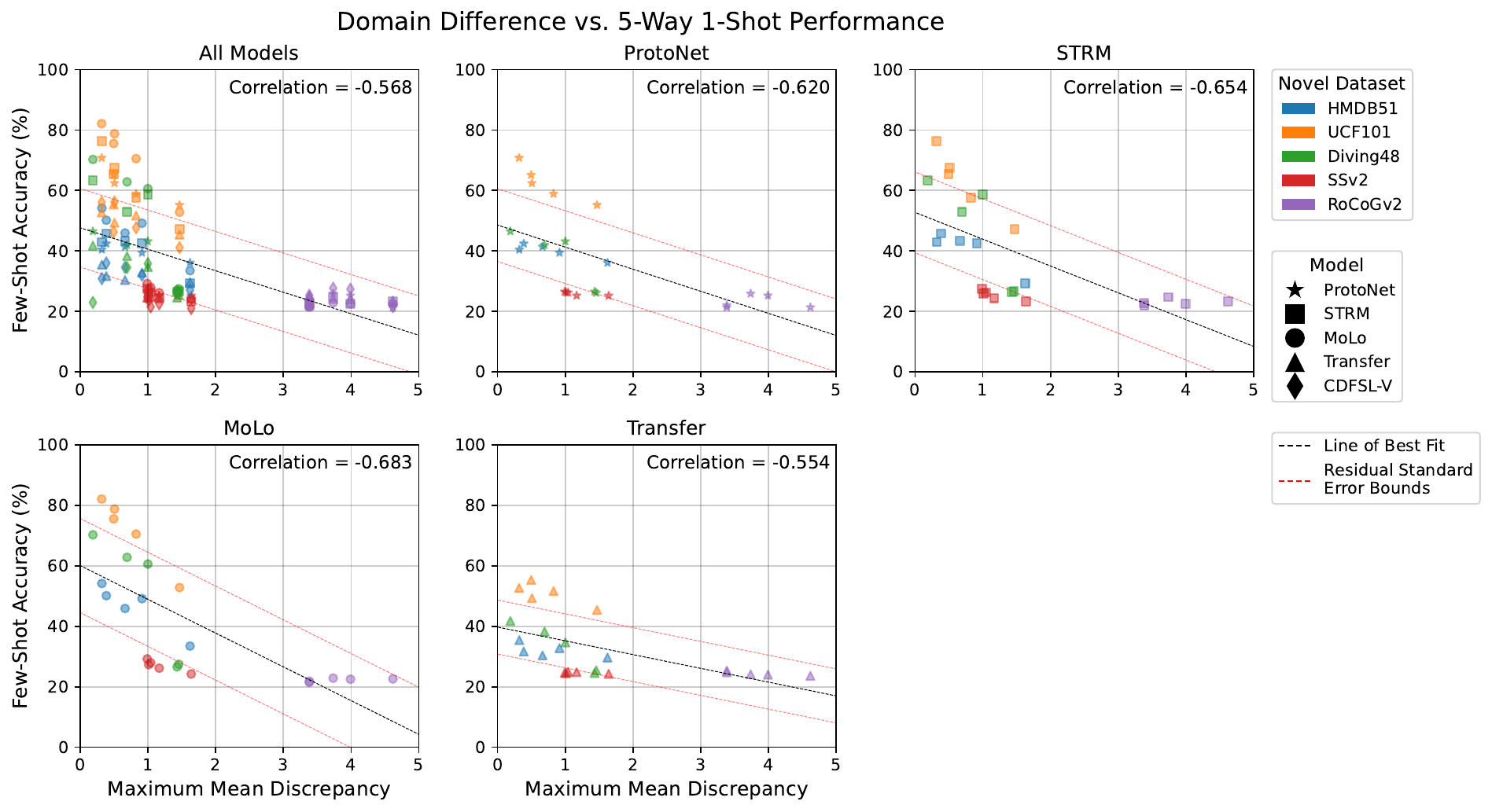}
	\caption{Measure of domain difference against downstream few-shot performance for each model under 5-Way 1-Shot evaluation conditions.}
	\label{fig:domain-diff-vs-fewshot-performance-1shot}
\end{figure}

See \Cref{fig:domain-diff-vs-fewshot-performance-1shot} for domain difference against downstream 5-way 1-shot performance. Again, the trends seen here are consistent with those analysed in the main body of the paper for 5-way 5-shot performance, where we witness a negative correlation between few-shot performance and domain difference for the single-domain method (ProtoNet, STRM, and MoLo), where as the domain difference decreases, the downstream few-shot performance increases. The transfer-based approach also follows this trend, but with a lower correlation.  

\end{document}